\documentclass{article}
\usepackage{log_2024}						

\usepackage{booktabs}						
\usepackage{multirow}						
\usepackage{amsfonts}						
\usepackage{graphicx}						
\usepackage{duckuments}						

\usepackage[numbers,compress,sort]{natbib}	

\usepackage[utf8]{inputenc} 
\usepackage[T1]{fontenc}    
\usepackage{url}            
\usepackage{booktabs}       
\usepackage{amsfonts}       
\usepackage{nicefrac}       
\usepackage{microtype}      
\usepackage{xcolor,colortbl}         
\usepackage{graphicx}
\usepackage{subfigure}
\usepackage{wrapfig}
\usepackage{capt-of}

\usepackage{multirow}           
\usepackage{duckuments}         

\usepackage{amsmath}
\usepackage{mathtools}
\usepackage{amsthm}
\usepackage{pifont}
\newcommand{\cmark}{\ding{51}}%
\newcommand{\xmark}{\ding{55}}%
\newcommand{\proj}{NLB\xspace}
\theoremstyle{plain}
\newtheorem{theorem}{Theorem}[section]

\theoremstyle{definition}
\newtheorem{definition}[theorem]{Definition}

\theoremstyle{remark}

\title[Scalable and Efficient Temporal Graph Representation Learning
via Forward Recent Sampling]{Scalable and Efficient Temporal Graph Representation Learning
via Forward Recent Sampling}

\author[Y. Luo and P. Li]{%
Yuhong Luo\\
Rutgers University\\
\email{y.luo@rutgers.edu}\And
Pan Li\\
Georgia Institute of Technology\\
\email{panli@gatech.edu}
}

\begin{document}

\maketitle

\begin{abstract}
Temporal graph representation learning (TGRL) is essential for modeling dynamic systems in real-world networks. However, traditional TGRL methods, despite their effectiveness, often face significant computational challenges and inference delays due to the inefficient sampling of temporal neighbors. Conventional sampling methods typically involve backtracking through the interaction history of each node. In this paper, we propose a novel TGRL framework, No-Looking-Back (NLB), which overcomes these challenges by introducing a forward recent sampling strategy. This strategy eliminates the need to backtrack through historical interactions by utilizing a GPU-executable, size-constrained hash table for each node. The hash table records a down-sampled set of recent interactions, enabling rapid query responses with minimal inference latency. The maintenance of this hash table is highly efficient, operating with $O(1)$ complexity. Fully compatible with GPU processing, NLB maximizes programmability, parallelism, and power efficiency. Empirical evaluations demonstrate that NLB not only matches or surpasses state-of-the-art methods in accuracy for tasks like link prediction and node classification across six real-world datasets but also achieves 1.32-4.40$\times$ faster training, 1.2-7.94$\times$ greater energy efficiency, and 1.63-12.95$\times$ lower inference latency compared to competitive baselines.

The link
to the code: \url{https://github.com/Graph-COM/NLB}.
\end{abstract}





 \vspace{-5mm}
\section{Introduction}\label{sec:intro}
 \vspace{-2mm}


Temporal networks are extensively employed to model real-world complex systems. Such networks dynamically evolve as interactions occur or states change within their nodes. Learning node representations in these networks is pivotal, as they can be used in many downstream tasks  such as  monitoring or predicting changes in the temporal networks. Representative applications include forecasting interactions between node pairs for the tasks such as anomalous transaction detection in financial networks~\cite{ranshous2015anomaly,wang2021bipartite,chang2020f}, recommendations in social networks~\cite{liben2007link} and in e-commerce systems~\cite{koren2009collaborative}. Node representations can also be used to predict individual nodes' properties such as community detection~\citep{Bhagat2011NodeCI, Kazemi2019RepresentationLF,liu2024deep} and  fraudsters detection~\cite{Wang2021One, Li2017Radar, Ding2019Deep, Yuan2021Higher}.



The interaction patterns surrounding a node often provide key indicators of its present state. Inspired by Graph Neural Networks (GNNs)~\cite{kipf2016semi, hamilton2017inductive} that may effectively encode the neighborhood of a node in static graphs, researchers have developed models for temporal networks to incorporate both the structural and temporal aspects of a node's historical neighborhood. These methods can be collectively named as temporal graph representation learning (TGRL)~\cite{tgn_icml_grl2020, wang2020inductive, xu2020inductive, luo2022neighborhood, kumar2019predicting, trivedi2019dyrep,wang2021apan}. 

Despite their effectiveness, previous TGRL methods generally require much more computational resources compared to their static graph counterparts. This is particularly challenging for large-scale applications that demand real-time inference. In such scenarios, reducing \emph{inference latency} — the time from query arrival to prediction response — becomes critical to avoid performance bottlenecks.

We identify that for most state-of-the-art (SOTA) TGRL methods, the online collection of the information around a node's neighborhood is a major computational bottleneck. As a query comes in, current methods have to backtrack
in time, sample a subset of historical interactions around the node of interest, and aggregate the information from these down-sampled neighboring interactions 
for prediction. Sampling neighboring nodes is a common strategy in representation learning for large static graphs~\citep{hamilton2017inductive,Ying2018Graph,Huang2018Adaptive,liu2023dspar}. However, this strategy faces challenges when applied to temporal networks, as they do not incorporate temporal information and may encounter high latency due to the enlarged sample space created by neighbors across various timestamps. 


\begin{wraptable}{R}{0.5\textwidth}
\centering
\vspace{-5mm}
\resizebox{0.5\textwidth}{!}{%
\begin{tabular}{r|cccc}
& forward sampling  & recent sampling & GPU sampling & node-level tasks \\
\hline
TGAT \footnotesize{\citep{xu2020inductive}} & \xmark &\cmark & \xmark & \cmark\\
TGN \footnotesize{\citep{tgn_icml_grl2020}} & \xmark &\xmark & \xmark & \cmark\\
CAWN \footnotesize{\citep{wang2020inductive}} & \xmark &\cmark & \xmark & \xmark\\
APAN \footnotesize{\citep{wang2021apan}} & \cmark &\xmark & \xmark & \cmark\\
NAT \footnotesize{\citep{luo2022neighborhood}} & \cmark &\xmark & \cmark & \xmark\\
\proj (this work) & \cmark &\cmark & \cmark & \cmark\\
\hline
\end{tabular}%
}
\caption{\small Comparisons between several TGRL methods.}
\label{tab:baselines}
\vspace{-3.3mm}
\end{wraptable}

To address this, a strategy termed \emph{recent sampling} has been proposed \citep{xu2020inductive,wang2020inductive}, where more recent interactions are given higher weights in sampling, such as through a probability proportional to $\exp(-c \triangle t)$, with $c\geq0$ and $\triangle t$ reflecting the time difference, indicating the recency of an interaction. While recent sampling can improve prediction performance, the process of calculating sampling weights and executing non-uniform sampling has high computational complexity $O(|\mathcal{N}_u^t|)$ where $|\mathcal{N}_u^t|$ denotes the size of the historical neighbors of node $u$ till time $t$. This significantly slows down the system. Consequently, most TGRL methods~\citep{tgn_icml_grl2020, wang2021apan, zhou2022tgl} opt for either simple \emph{uniform sampling} or a strategy named \emph{truncation}, where only a fixed number of most recent neighbors are considered, to reduce the complexity. Even with such simplification, the inference latency may still be high because backtracking and sampling of these historical interactions are performed within CPUs instead of GPUs due to the inherent irregularity. A recent framework called TGL~\citep{zhou2022tgl} reduced the issue via leveraging powerful multi-core CPUs complemented by meticulously crafted C++ parallelizable programming for fine-grained memory and thread management.
%
%
However, multi-threading in CPUs is less power-efficient and may still suffer from longer latency than parallelism in GPUs.

In this paper, we propose \emph{\textbf{N}o-\textbf{L}ooking-\textbf{B}ack} (\proj), an efficient and scalable framework for TGRL that aims to improve training and inference latency, and power efficiency without compromising on prediction accuracy. It adopts a novel sampling strategy called \emph{forward recent sampling} that allows getting around  
backtracking historical interactions while achieving the benefits of recent sampling. We compare \proj with existing methods in Table~\ref{tab:baselines}. Our key idea is to maintain a GPU-executable size-constrained hash table for each node that records a set of down-sampled recent interactions.  These down-sampled interactions can be directly used to track neighbors and to generate response for upcoming queries with extremely low inference latency. Moreover, the maintenance of this table only requires $O(1)$ complexity to deal with the new link. Specifically, we leverage hash collision to insert the new link into this table randomly to simulate the strategy of recent sampling. Importantly, all of these operations are implementable with PyTorch and are fully compatible with GPU processing. This ensures not only maximized programmability and parallelism but also improved power efficiency.

\proj supports two variants \proj-edge and \proj-node, which differ in the used hash keys being either link ID or node ID. \proj-edge and \proj-node provably achieve recent sampling of links and recent sampling of nodes respectively while both support node-level and link-level applications. Empirically, \proj is comparable or improves SOTA link prediction and node classification accuracy in 6 real-world datasets while being 1.32-4.40$\times$ faster in training and 1.2-7.94$\times$ more energy efficient than the three most competitive baselines, and 1.63-12.95$\times$ more efficient in inference latency than the fastest baseline that leverages powerful multi-thread CPUs to perform backward sampling.
\begin{figure*}[t]
    \includegraphics[trim={0.3cm 0.3cm 0.3cm 0.4cm}, clip, width=1.0\textwidth]{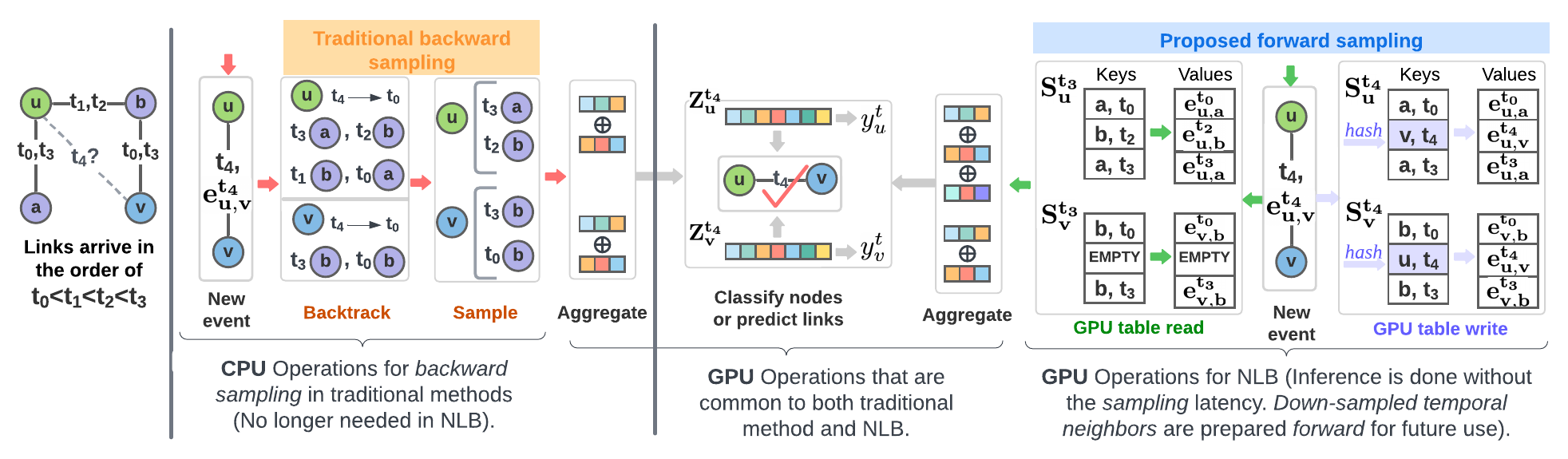}
    \vspace{-0.4cm}
    \caption{\small{A toy example comparing the sampling done in current TGRL methods v.s.~in NLB. \emph{Left}: Given the toy historical temporal network, the task is either (1) to classify nodes $u$ and $v$ at time $t_4$, or (2) to predict whether $u$ will connect with $v$ at $t_4$. \emph{Middle}:   As a new link $(u,v,t_4)$ arrives, traditional methods require looking backward and sampling a subset of historical temporal neighbors (via CPU). The generated embeddings of the temporal neighbors are aggregated to generate the final representations $Z_u^{t_4}$ and $Z_v^{t_4}$ for making predictions. \emph{Right}: \proj abandons the backward sampling in traditional methods and adopts forward sampling. For inference, it directly looks up the down-sampled temporal neighbors from GPU hash tables for aggregation and prediction without being slowed down by the sampling operations. The new interaction can replace older temporal neighbors in the hash tables leveraging hash collision with complexity $O(1)$. The updated hash tables capture the new down-sampled temporal neighbors for use later on.}}
    \label{fig:alg}
    \vspace{-5mm}
\end{figure*}

\vspace{-4.5mm}
\section{More Related Works}
\label{sec:relatedwork}
\vspace{-2mm}

There are two streams of work on representation learning for temporal networks. Earlier work proposed aggregating the sequence of links into network snapshots and processing a sequence of static network snapshots~\citep{zhou2018dynamic,du2018dynamic,mahdavi2018dynnode2vec,singer2019node,nguyen2018continuous, sankar2020dysat}, but they may suffer from low prediction
accuracy as they cannot capture time-sensitive information within static snapshots. More recent temporal graph learning methods process link streams directly~\citep{trivedi2017know,trivedi2019dyrep,kumar2019predicting,souza2022provably, Poursafaei2022Towards, cong2023do}. Most of these methods learn representation for a node using the historical information from the neighboring nodes it previously interacted with~\citep{xu2020inductive,tgn_icml_grl2020,wang2020inductive, zhou2022tgl,luo2022neighborhood,yu2023towards,besta2023hot, tian2024freedyg}. 


There are many works for graph learning that consider improvement on scalability and inference efficiency. For GNNs on static graphs, a stream of work focuses on efficient neighbor sampling~\citep{hamilton2017inductive,Ying2018Graph,Huang2018Adaptive,liu2023dspar} which speeds up both training and inference, while some other works studies the acceleration for inference via knowledge distillation~\citep{zhang2022graphless,guo2023linkless,Wang2023graph} or channel pruning~\citep{zhou2021accelarating}. However, these techniques cannot be easily generalized to improve the scalability for temporal networks. First, as temporal links come in as a sequence, it is non-scalable to construct and maintain the whole graph for every new link. Furthermore, these methods do not resolve the bottleneck of historical neighbor sampling on temporal graphs, but neighborhood information is crucial for graph representation learning.

For temporal graphs, due to the more significant computation overhead of TGRLs as reviewed in Sec.~\ref{sec:intro}, several recent works focusing on acceleration and efficiency inference are proposed~\citep{wang2021apan,zhou2022tgl,luo2022neighborhood, Zhou2023DistTGL}. APAN~\citep{wang2021apan} proposed asynchronous graph propagation which passes information to the most recent temporal neighbors and works with a sampling property similar to truncation without sampling. It improves inference latency but requires significant memory for buffering messages and the recent truncation strategy may hurt prediction accuracy. A recent work NAT~\citep{luo2022neighborhood}  aims to improve link prediction performance and scalability by learning neighborhood-aware representations, which elegantly avoids backward sampling. However, its sampling property is not mathematically explained, and it relies on constructing joint neighborhood structural features of node pairs, which cannot be generalized to node-level tasks. 
In experiment (Sec.~\ref{sec:exp}), we find that by removing the joint neighborhood features from NAT, it can be applied to node-level tasks. But its node-level prediction performance is worse than \proj and it is still slow compared to \proj. 

Finally, temporal graph benchmark (TGB)~\citep{huang2023temporal} is proposed to standardize the evaluation pipelines for TGRL. We build a separate evaluation pipeline outside of TGB because our focus involves computational evaluations on billion-scale datasets which has not been supported by TGB. We provide supplementary evaluation of \proj on TGB in Appendix~\ref{apx:tgb}.
\vspace{-4mm}
\section{Notations and Problem Formulation}
\vspace{-1.5mm}

\label{sec:prelim}

Next, we introduce some notations and the problem formulation. We consider temporal network as a sequence of timestamped interactions between pairs of nodes.
\begin{definition}
    [Temporal network] A temporal network $\mathcal{E}$ can be represented as 
 $\mathcal{E} = \{(u_1, v_1, t_1), (u_2, v_2, t_2),\cdots\}$, $t_1 \le t_2 \le \cdots $ where $u_i$ and $v_i$ denote interacting node IDs of the $i$th  link, $t_i$ denotes the timestamp. Each temporal link $(u,v,t)$ may have link features 
 $e_{u,v}^t$. We also denote the entire node set as $\mathcal{V}.$ Without loss of generality, we use integers as node IDs, i.e., $\mathcal{V} = \{1,2,...\}$. For simplicity of the notations, we assume the links are undirected. 
 \end{definition}
The dynamics of a node's neighborhood provide crucial information about a node current state. We define the temporal neighbors of a node as follows.
\begin{definition} [Temporal neighbors of a node] We denote the temporal neighbors of node $u$ at time $t$ as all neighbors interacted in the past. They are represented by tuples of opposing node IDs, link features and timestamps: \begin{align}\label{eq:neighbor}
    \mathcal{N}^t_u \coloneqq \{ (v, e^{t'}_{u,v}, t')\,\,|\,\,(u, v, t') \in\mathcal{E}, t'<t \}.
\end{align}
    
\end{definition}

Finally, we formulate our problem as follows. 
\begin{definition} [Problem formulation]  Our problem is to learn a model that uses the historical information including the temporal neighbors about a node (e.g., $u$) before a timestamp (e.g., $t$) 
to efficiently generate a node representation that is predictive for downstream tasks including link prediction and node classification.  We denote the representation at timestamp $t$ for each node $u \in \mathcal{V}$ as $Z_u^t$. We define the link prediction and node classification problems as follows. 

\emph{Link prediction:} given any pair of representations, $(Z_u^t, Z_v^t)$, a link prediction task predicts whether there will be a temporal link between the two nodes at time $t$, i.e., $(u,v,t) \in \mathcal{E}$.
\emph{Node classification:} a node $u$ can be dynamically labeled, Let $y_u^t$ denote the label of $u$ at timestamp $t$. A node classification task predicts $y_u^t$ given the representation $Z_u^t$. 
\end{definition}
\vspace{-4.5mm}
\section{Methodology}\label{sec:method}
\vspace{-2mm}


In this section, we first review  the sampling strategies of previous
approaches. We then introduce forward recent sampling, the sampling strategy of \proj and show that it provably achieves recent sampling. Lastly, we present the efficient aggregation of temporal neighbors and the generation of node presentations for inference. 

\vspace{-2.5mm}
\subsection{The sampling strategies of existing works} 
\vspace{-1.5mm}
 While different approaches have different implementations, a common computational bottleneck for generating node representations during both training and inference is the online sampling of temporal neighbors. Existing sampling strategies consist of two steps: (1) Backtrack and collect the historical interactions of a node from the current timestamp; (2) Sample a subset of temporal neighbors from those interactions. Traditional methods~\citep{xu2020inductive, tgn_icml_grl2020,wang2020inductive, zhou2022tgl}  perform these operations within CPUs, causing inefficiency in both time and power consumption for both training and inference. 


To give an example, to learn the representations for node $u$ and $v$ at $t_4$ for downstream predictions  (Fig.~\ref{fig:alg} Left),  traditional methods   perform backtracking and sampling within the CPUs (Fig.~\ref{fig:alg} Middle). These blocking operations not only slow down the training and inference processes, but also introduce costly energy consumption from both CPU computation and CPU-GPU communication. 
Previous methods commonly adopted the following sampling strategies.

\begin{definition}[Truncation]
    Truncation of a fixed number of most recent neighbors, or simply named \emph{truncation}, is a common strategy used by TGN~\citep{tgn_icml_grl2020}, APAN~\citep{wang2021apan} and TGL~\citep{zhou2022tgl}. It does not sample with randomness but truncates the most recent $s$ temporal neighbors that appear in the history instead. Truncation is effective when the most recent neighbors provide sufficient information about the current state of a node's neighborhood. However, as it does not allow randomness, it may fail to capture information that appears earlier in the history. In an extreme, if all most recent interactions are from the same neighbor, truncation neglects information from all other early neighbors.
\end{definition}

\begin{definition}[Uniform sampling]
\emph{Uniform sampling} is also commonly supported by TGN, TGL, TGAT~\citep{xu2020inductive}, etc. 
It samples among all temporal neighbors in the past with equal probabilities. It is effective in capturing
early-day neighbors which may benefit prediction. However, for many real-world applications, more recent events typically play a more important role. Moreover,  uniform sampling, albeit straightforward, still requires the access to all of the historical temporal interactions. Since different nodes have different numbers of historical interactions, uniform sampling can hardly be implemented in the GPUs for parallelism, and thus is slower than truncation.
\end{definition}

\vspace{-2.9mm}
\subsection{Our approach: Forward recent sampling }\label{sec:sampling}
\vspace{-1.5mm}

In this section, we will  introduce recent sampling, then propose forward sampling that implements recent sampling efficiently in $O(1)$ and provide theorectical analysis in Sec.~\ref{sec:theory}. 

\begin{definition}[Recent sampling]\label{def:recent} 
Recent sampling assigns larger sampling weights to more recent temporal neighbors. The probability a temporal neighbor of node $u$, e.g., $(v, e^{t'}_{u,v}, t')$ is sampled for learning $Z_u^t$ is proportional to $\exp(c(t' - t))$ where $c \geq 0$ is a constant.
This sampling strategy is previously considered in \citet{xu2020inductive} and \citet{wang2020inductive}.
Recent sampling has shown to be effective for many scenarios as both recent and some early-day  neighbors could provide information. By tuning $c$, recent sampling may approximate uniform sampling (for a small $c$) and truncation (for a large $c$). 
However, it is hard to implement efficiently via the traditional backward sampling. Not only the existing way to implement recent sampling needs to access all the timestamps of historical interactions to compute those non-uniform sampling probabilities, but also performing non-uniform sampling is rather irregular, which could be much slower than truncation and uniform sampling. 
\end{definition}

\textbf{Forward Recent Sampling.} For both uniform sampling and recent sampling, if they are achieved through backward sampling as practiced by current methods, the time complexity is at least $O(|N_u^t|)$ as they need to access the entire historical neighbors. We now introduce forward sampling that implements recent sampling in $O(1)$. An overview of the algorithm is demonstrated in Fig.~\ref{fig:alg} Right. 

Intuitively, to abandon the expensive backtracking and sampling procedures, we adopt a dictionary-type table that keeps track of a list of down-sampled temporal neighbors for each node. The storage tables are updated after the model has responded to the previous query but before the next query comes in. So, the update will not introduce inference latency, although the update itself is cheap (of $O(1)$ complexity)  and parallelizable on GPUs.  

Formally, we denote $S_u$ to be the down-sampled temporal neighbors of node $u$ stored within a GPU hash table. For efficient lookup and update in batches, we allocate fixed-size GPU memory for these tables, where the size $s$, as a user-specified parameter, is the maximum number of temporal neighbors to be sampled. The current snapshot of table $S_u$ at a timestamp $t$ is denoted as $S_u^t$. $S_u$ consists of key-value pairs with unique keys. Each key-value pair corresponds to one of the temporal neighbors of $u$, e.g., $(v,e^{t'}_{u,v}, t') \in \mathcal{N}_u^t$. While we store the edge features 
$e^{t'}_{u,v}$ as the \emph{values}, the choice of the \emph{keys} is flexible: It can either be the pair of the neighbor ID and timestamp, i.e., ~$(v, t')$, or just the neighbor ID $v$. 
We name the method using $(v, t')$ as \emph{\proj-edge} and the one using $v$ as \emph{\proj-node}. Different choices of the keys have different sampling properties 
which we will discuss in Sec.~\ref{sec:theory}.

To achieve recent sampling, the tables get updated overtime when  new links come in. 
Specifically, the new temporal neighbor indicated by the new link is inserted into the table through random hashing using the keys. When there is a hash collision,  the older temporal neighbor will be replaced by the new one with a probability $\alpha\in(0,1]$, where $\alpha$ is a hyperparameter. In order for more recent neighbors to be sampled, we set $\alpha$ to be closer to 1.

\textbf{Example.} In Fig.~\ref{fig:alg}, when link $(u,v,t_4)$ arrives, we initialize an update to $S_u$ and $S_v$, creating new snapshots $S_u^{t_4}$ and $S_v^{t_4}$. 
Specifically, for \proj-edge which uses tuples of neighbor ID and timestamp as keys, the new temporal neighbor $(v, e^{t_4}_{u,v}, t_4)$ of $u$ will be assigned to position 
\begin{align}
    \emph{hash}(v,t_4) \equiv (q_1*v + q_2*t_4) \;(\textbf{mod}\,s) \label{eq:hash_e}
\end{align}
for fixed large prime numbers $q_1$ and $q_2$. For \proj-node which uses neighbor IDs as keys, the temporal neighbor will be assigned to position \begin{align}
    \emph{hash}(v) \equiv (q_1*v) \;(\textbf{mod}\,s). \label{eq:hash_n}
\end{align}
Let $a$ denote the assigned position for $(u,v,t_4)$. Then,  $(v, e^{t_4}_{u,v}, t_4)$ will be inserted into $S_u$, if there is no hash collision, i.e., $S_u[a]$ being empty. Otherwise, the temporal neighbor will be inserted with a probability $\alpha$.


This completes the procedure for  forward sampling. At future timestamp $t > t_4$, $S_u^{t_4}$ and $S_v^{t_4}$ can be used as the down-sampled temporal neighbors for learning node representations $Z_u^t$ and $Z_u^t$ until new snapshots of $S_u$ and $S_v$ are generated. How \proj encodes these temporal neighbors into node representations will be discussed in Sec.~\ref{sec:rep_gen}.

\vspace{-3mm}
\subsection{The properties of forward recent sampling}
\label{sec:theory}
\vspace{-2mm}
The sampling properties for different choices of the keys are different. Intuitively, a key with both neighbor IDs and timestamps allows \proj-edge to keep track of more dynamic information about each neighboring node, while hashing with only neighbor IDs allows \proj-node to learn more static information. 
In the following Theorem~\ref{thm:edge}, we show that \proj-edge essentially implements recent sampling (Def.~\ref{def:recent}) with $O(1)$ complexity. The proof can be found in Appendix~\ref{apx:nlbedgeproof}. \proj-node actually implements a node-wise recent sampling rather than the standard recent sampling with $O(1)$ complexity, which will be elaborated in Appendix~\ref{apx:nlb-node}.

\begin{theorem}[\proj-edge achieves recent sampling]\label{thm:edge}  
Suppose links come in for any node (e.g.~$u$) by following a Poisson point process with a constant intensity (e.g.~$\lambda$), and suppose a temporal neighbor $(v, e_{u,v}^{t_i},t_i)$ is inserted into $S_u$ at time $t_i$, then $\Pr((v,e_{u,v}^{t_i},t_i) \in S_u^t) = \exp(\frac{\alpha\lambda}{s}(t_i - t))$.
\footnote{Poisson point process is a commonly
used assumption to model communication networks~\citet{lavenberg1983computer}, and is also an assumption used by CAWN~\citep{wang2020inductive}.} 
\end{theorem}

This result matches the definition of recent sampling (Def.~\ref{def:recent}) where the probability a temporal neighbor gets sampled is proportional to $\exp(c(t_i - t))$. If we instead suppose that each unique neighboring node (e.g.~$v$) of $u$ connects with $u$ following a different Poisson point process (e.g.~$\lambda_v$), we recover the sampling probabilities of \proj-node.

\vspace{-2mm}
\subsection{Node representation generation and \proj prediction}\label{sec:rep_gen}
\vspace{-2mm}

When a prediction query comes,  \proj generates the node representations based on the sets of down-sampled temporal neighbors of relevant nodes, as introduced in Sec.~\ref{sec:sampling}, for such prediction. 
We first denote the status of node $u$ at time $t$ as $r_u^t$, which can be viewed as a self feature vector aggregated over time via RNN and will be elaborated later.
Node status can be viewed as to collect local information, while the node representation $Z_u^t$ also leverages the status of the temporal neighbors (decided by $S_u^t$), which can be viewed as to provide more global information. Empirically, we observe that combining both yields the best performance.


\textbf{Prediction.} 
Given the down-sampled neighbors $S^t_u$ of $u$, we first retrieve the status of the temporal neighbors $\{r_v^t\}$. We then use attention to aggregate the collected status $\{r_v^t\}$ and the self status $r_u^t$. The timestamps and edge features for temporal neighbors will also be encoded together. 
Specifically, let MLP denote a multi-layer perceptron. We adopt
\begin{align} \label{eq:dgat}
Z^t_u = \text{MLP}\bigg(r_u^t, \sum_{(v, e^{t'}_{u,v},t')\in S^t_u} \alpha_i \text{MLP}(r_v^t, e^{t'}_{u,v}, t')\bigg),\end{align} 
$
\text{where}\,\{\alpha_i\} = \text{\small{softmax}}(\{w^T\text{MLP}(r_v^t, e^{t'}_{u,v}, t')|(v, e^{t'}_{u,v},t')\in S^t_u\})$, 
where $w$ is a learnable vector parameter. A pair of node representations $Z^t_u$ and $Z^t_v$ can be combined and plugged in to classifiers to make link predictions; and a single node representation $Z^t_u$ can be used for node classification. $t$ is encoded via T-encoding (see Def.~\ref{def:tencoding} in Appendix~\ref{apd:tencoding}) which has been proved to be effective in previous works~\cite{xu2019self,kazemi2019time2vec,xu2020inductive, wang2020inductive}.

Note that a node status $r_u^t$ is updated via  $r_{u}^{t+1}\leftarrow \textbf{RNN}([r_{u}^t, r_v^t, t, e^t_{u,v}])$ after an event $(u,v,t)$ arrives.

Also, \proj naturally supports efficient aggregation of second-hop temporal neighbors because to get representation for node $u$, if $v$ is stored in $S_u^t$, i.e., one of the first-hop temporal neighbors, $S_v^t$ provides the down-sampled second-hop neighbors of $u$, which can be efficiently accessed. 
However, in experiment, we use the first hop only as it already provides decent prediction performance. Involving the second hop, albeit bringing more information, will increase the inference latency. 




 \vspace{-3mm}
\section{Experiments}
\label{sec:exp} 
\vspace{-2mm}



%
    

In this section, we evaluate \proj in its prediction performance and scalability 
on real-world temporal networks, and further conduct hyperparameter analysis. 
\vspace{-2mm}
\subsection{Experimental setup}\label{sec:exp_setup}
\vspace{-1mm}

\textbf{Datasets.} We use six publicly available real-world datasets whose statistics are listed
in Appendix Table~\ref{tab:dataset} for experiments. There are datasets with billion-scale temporal links and million-scale nodes. Further details of these datasets can be found in Appendix~\ref{apd:dataset}.
We split the datasets into training, validation and testing data according to the ratio of 70/15/15 while preserving the chronological order.

\textbf{Downstream tasks and models.} 
 We conduct experiments on link prediction with transductive and inductive settings for all datasets. For datasets where node labels are available, we also conduct node classification. For inductive link prediction, we follow previous work and sample the unique nodes in validation and testing data with probability 0.1 and
remove them and their associated edges from the networks during the training stage. 
Following previous work, the temporal graph representation models are trained with self-supervised link prediction task. The models are then  directly used to generate node representations for node classification. For fair comparison, the architectures for all downstream models are single-layer perceptrons with ReLU activation. The detailed
procedures for inductive evaluation and node classification for \proj are documented in Appendix~\ref{apd:hyper}.

\textbf{Baselines.} We select four representative baselines, each with two variants. Among the baselines, 
TGAT~\citep{xu2020inductive} proposed aggregating temporal neighborhood information with attention, TGN~\citep{tgn_icml_grl2020} proposed keeping a memory state for each node that gets update with new interactions, APAN~\citep{wang2021apan} proposed asynchronous graph propagation instead of graph aggregation, and NAT~\citep{luo2022neighborhood} improves link prediction with their joint neighborhood structural features. For TGAT, TGN and APAN, we evaluate their performance and scalability with two sampling strategies: uniform sampling (denoted as \emph{unif}) and truncation of the most recent neighbors (denoted as \emph{trunc}). We evaluate these baselines (excluding NAT) using the implementation in TGL~\citep{zhou2022tgl} that relies on low-level C++ programming and multi-thread CPU, as it is currently the most efficient framework for them and the original implementation has shown to be significantly slower. For NAT, we adapt it for node classification (named \emph{NAT-node}) by removing their joint structural features and directly aggregating their neighborhood representations to generate node representations. Additional details about these
baselines can be found in Appendix~\ref{apd:hyper}.

\textbf{Hyperparameters.} For fair comparison, we fix the maximum number of neighbors to be sampled 
for each dataset across all methods as specified in Appendix Table~\ref{tab:hyperparam}.  When evaluating scalability, we (1) limit the maximum number of CPU threads to be at most 32, (2) use one GPU for all methods and (3) fix the batch sizes across different methods as specified in Appendix Table~\ref{tab:bs}. For the rest of the hyperparameters, if a dataset has been tested previously, we use the set of hyperparameters that are provided by the baseline models. Otherwise, we tune the parameters with grid search and make sure the sizes of different modules such as the node representation and time feature, have  dimensions of the same scale for all methods. More detailed hyperparameters are provided in Appendix~\ref{apd:hyper}.
\textbf{Hardwares.} We run all experiments using the same Linux device that is equipped with 256 AMD EPYC 7763 64-Core Processor @ 2.44GHz with 2038GiB RAM and one GPU (NVIDIA RTX A6000).



\textbf{Performance Metrics.} For link prediction performance, we evaluation all models with Area Under the ROC curve (AUC), Average Precision
(AP) and Mean Reciprocal Rank (MRR) with a large number of negative samples per positive sample (50 for GDELT, and 500 for other datasets, excluding the MAG dataset due to the time and memory constraints). In the main text, the prediction performance in all tables
is evaluated in AUC. 
For node classification, we evaluate with AUC for Wikipedia and Reddit which have two node classes, and we use F1 for GDELT and MAG which have more than two classes. 
All results are summarized based on 5
time independent experiments with different random seeds and initializations.

\textbf{Scalability Metrics.} To evaluate scalability, we consider both time and energy efficiency. We include (a) average training time (Train), testing time (Test) and inference latency (Inf.~Lat.) per epoch in seconds, (b) the total energy consumption from CPUs and GPUs (in Joules) denoted as CPU and GPU.
We ensure that there are no other applications running during our evaluations.

\begin{table*}[t]
    \centering     
\resizebox{0.8\textwidth}{!}{%
    \begin{tabular}{c|r|cccccc}
    \hline
    \multicolumn{1}{c|}{Task} & Method & \multicolumn{1}{c}{Wikipedia} & \multicolumn{1}{c}{Reddit}& \multicolumn{1}{c}{GDELT}  & \multicolumn{1}{c}{MAG} & \multicolumn{1}{c}{Ubuntu} & \multicolumn{1}{c}{Wiki-talk}\\
    \hline
    \multicolumn{1}{c|}{\multirow{10}{*}{\rotatebox[origin=c]{90}
    {Transductive}}}
    & TGN-trunc & \underline{99.48 $\pm$ 0.07} & {99.65 $\pm$ 0.07} & \underline{98.63 $\pm$ 0.05} & \underline{99.32 $\pm$ 0.06}  & 76.09 $\pm$ 0.17 & {84.73 $\pm$ 3.57}\\
     & TGN-unif & \underline{99.44 $\pm$ 0.03} & {99.66 $\pm$ 0.05} &  97.85 $\pm$ 0.02  & {99.24 $\pm$ 0.03}  &{78.99 $\pm$  1.69} & {83.61 $\pm$  0.10}\\
      & TGAT-trunc & {97.47 $\pm$ 0.06} & {97.84 $\pm$ 0.03} & {98.31 $\pm$ 0.02} & { 99.09 $\pm$ 0.03}  & 76.71 $\pm$ 0.33 &  
 {81.23 $\pm$ 0.06}\\
       & TGAT-unif & {95.35 $\pm$ 0.18} & {98.15 $\pm$ 0.15} &  97.76 $\pm$ 0.01 & {99.02 $\pm$ 0.05}  & {78.59 $\pm$ 0.23} & 80.48 $\pm$ 0.01\\
        & APAN-trunc & {99.20 $\pm$ 0.03} & {96.46 $\pm$ 2.98} & 97.89 $\pm$ 0.16 & {90.61 $\pm$ 1.24}  & 69.90 $\pm$ 3.28  & {82.10 $\pm$ 5.67}\\
         & APAN-unif & {97.90 $\pm$ 0.40} & {97.65 $\pm$ 0.20} & 97.56 $\pm$ 0.74 & CPU OOM  & {77.62 $\pm$ 2.71} & {84.32 $\pm$ 7.28}\\
            & NAT & \textbf{99.72 $\pm$ 0.03} & \textbf{99.90 $\pm$ 0.01} & GPU OOM  & GPU OOM  &  \textbf{89.65 $\pm$  0.17} & \textbf{94.66 $\pm$ 0.03} \\
        & NAT-node & {99.29 $\pm$ 0.07} & \underline{99.76  $\pm$ 0.02 } & {GPU OOM} & {GPU OOM} & \underline{87.48 $\pm$ 0.49} & \underline{93.03 $\pm$ 0.52}\\
     \multicolumn{1}{c|}{} & \textbf{\proj-edge} & \underline{99.42 $\pm$ 0.08} & {99.73 $\pm$ 0.02} & \textbf{98.94 $\pm$ 0.28} & \textbf{99.39  $\pm$ 0.02}  & \underline{87.18 $\pm$ 0.55}  & {91.72 $\pm$ 0.62} \\
     \multicolumn{1}{c|}{} & \textbf{\proj-node} & {99.03 $\pm$ 0.07} & {99.67 $\pm$ 0.02} & \textbf{98.80 $\pm$ 0.16} & {99.15 $\pm$ 0.02}  & \underline{87.48 $\pm$ 0.64}  & {91.95  $\pm$ 0.17} \\
     \hline
    \multicolumn{1}{c|}{\multirow{10}{*}{\rotatebox[origin=c]{90}
    {Inductive}}}
    & TGN-trunc &  \underline{98.55 $\pm$ 0.08} & \underline{99.44 $\pm$ 0.06} & \underline{98.62 $\pm$ 0.01} & {96.05 $\pm$ 0.12}  &  80.70 $\pm$ 1.64 & 87.65 $\pm$ 1.01\\
     & TGN-unif &  {98.33 $\pm$ 0.08} & {99.30 $\pm$ 0.01} & \textbf{98.77 $\pm$ 0.08} & 96.45 $\pm$ 0.26  &  \textbf{85.23 $\pm$ 1.66} & {87.11 $\pm$ 0.30}\\
  & TGAT-trunc &  {96.57 $\pm$ 0.04} & {95.22 $\pm$ 0.21} & 94.36 $\pm$ 0.01 & \underline{98.80 $\pm$ 0.01}  &  {77.22 $\pm$ 0.07} & 79.97 $\pm$ 0.04\\
   & TGAT-unif &  {93.80 $\pm$ 0.15} & {96.12 $\pm$ 0.07} & 94.05 $\pm$ 0.02 &  {98.77 $\pm$ 0.01} &  78.07 $\pm$ 0.04  & 77.03 $\pm$ 0.03\\
    & APAN-trunc &  {96.65 $\pm$ 0.07} & {95.93 $\pm$ 2.43} & 98.48 $\pm$ 0.01 & {CPU OOM}  & 65.19 $\pm$ 5.50 & 78.90 $\pm$ 1.11\\
     & APAN-unif &  {97.23 $\pm$ 0.31} & {96.56 $\pm$ 0.32} & 97.64 $\pm$ 0.14 &  CPU OOM & 78.39 $\pm$ 2.87 & 87.86 $\pm$ 0.11\\
        & NAT & \textbf{99.52 $\pm$ 0.03} & \textbf{99.78 $\pm$ 0.03} & GPU OOM & GPU OOM & \underline{84.71 $\pm$ 1.01} &  \textbf{92.23 $\pm$ 1.16}\\
        & NAT-node & \underline{98.47 $\pm$ 0.25} & \underline{99.53 $\pm$ 0.21 } & {GPU OOM} & {GPU OOM} & {81.51 $\pm$ 1.36} & {79.12 $\pm$ 5.14}\\
     \multicolumn{1}{c|}{} & \textbf{\proj-edge} &  \underline{98.43 $\pm$ 0.26} & \underline{99.43 $\pm$ 0.07} & 98.16 $\pm$ 0.32 & \textbf{98.85 $\pm$ 0.02}  & \textbf{86.88 $\pm$ 0.62} & \underline{90.91 $\pm$ 1.14} \\
     \multicolumn{1}{c|}{} & \textbf{\proj-node} &  \underline{98.31 $\pm$ 0.30} & \underline{99.36 $\pm$ 0.09} & 97.14 $\pm$ 0.42 & \textbf{98.79 $\pm$ 0.11}  &  
 \underline{85.47 $\pm$ 0.73}  & \textbf{91.22 $\pm$ 1.39} \\
    \hline
    \end{tabular}
    }
    \vspace{-1mm}
    \caption{\small Link prediction performance in  AUC (mean in percentage $\pm$ 95$\%$ confidence level). \textbf{Bold font} and {underline} highlight the best performance and the second best performance on average.} 
    \label{tab:auc results}
\vspace{-4mm}
\end{table*}
\begin{figure}
\begin{minipage}{0.37\textwidth}
\centering 
\includegraphics[trim={0.3cm 0.3cm 0.3cm 1.3cm}, clip, width=\textwidth]{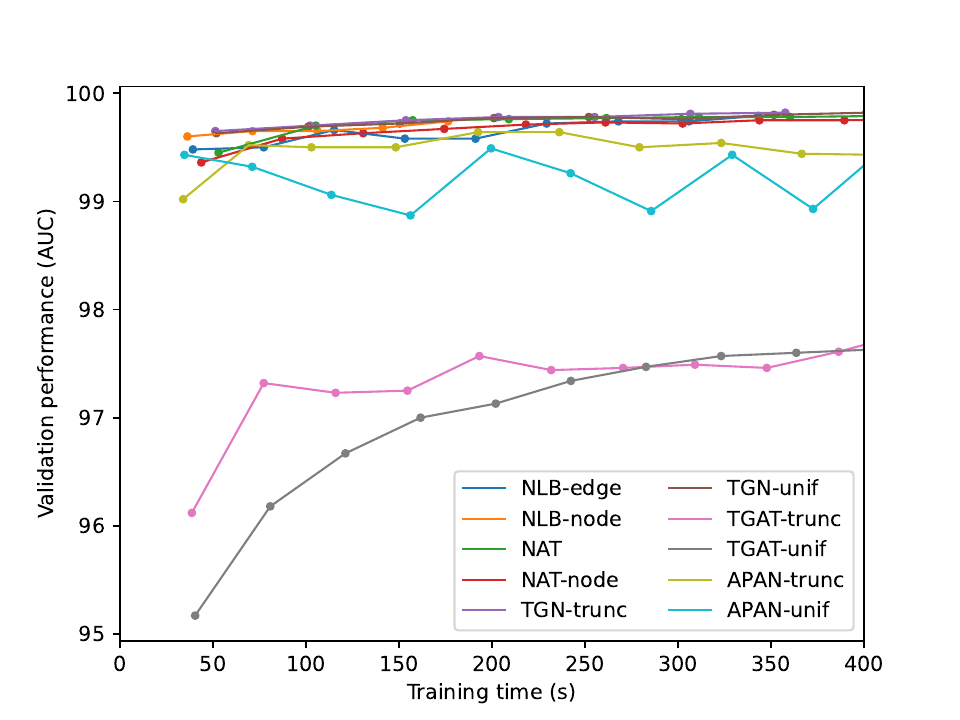}
    \vspace{-5mm}
    \caption{\small{The transductive link prediction validation performance v.s.~training time on Reddit. Each dot on the curves gets
collected at the end of an epoch.}}
    \label{fig:converge}
\end{minipage}
\hfill
\begin{minipage}{0.6\textwidth}
    \centering
    \resizebox{0.99\textwidth}{!}{%
    \begin{tabular}{r|cccccccc}
    \hline
    \multicolumn{1}{r}{Method}  & \multicolumn{1}{c}{Wikipedia (AUC)} & \multicolumn{1}{c}{Reddit (AUC)}&\multicolumn{1}{c}{GDELT (F1)} & \multicolumn{1}{c}{MAG (F1)}\\
    \hline
    
     TGN-trunc &  {86.85 $\pm$ 0.56} &  \underline{63.45 $\pm$ 1.21} & 18.39 $\pm$ 1.37 &2.33 $\pm$ 0.00\\
      TGN-unif & {85.99 $\pm$ 0.62} &  \textbf{64.38 $\pm$ 1.60} & 21.38 $\pm$ 0.04 & 2.33 $\pm$ 0.00\\
      TGAT-trunc & {78.27 $\pm$ 1.12} &  {60.54 $\pm$ 0.97} & 22.91 $\pm$ 0.03  & 2.33 $\pm$ 0.00\\
     TGAT-unif & {84.08 $\pm$ 0.41} &  \underline{63.44 $\pm$ 2.72} &  \textbf{24.34 $\pm$ 0.02} & 2.33 $\pm$ 0.00\\
      APAN-trunc & {86.50 $\pm$ 0.60} &  {56.74 $\pm$ 0.62} & 9.14 $\pm$ 0.43 & 3.11 $\pm$ 0.00\\
      APAN-unif & {87.27 $\pm$ 1.02} &  {60.68 $\pm$ 1.42} &  
9.90 $\pm$ 0.01 & CPU OOM \\
      NAT & - &  - & - & -\\
    NAT-node & {86.65 $\pm$ 0.55} & {60.44 $\pm$ 2.36} & {GPU OOM} & {GPU OOM} \\
     \textbf{\proj-edge} & \textbf{89.92 $\pm$ 0.20} &  \underline{62.73 $\pm$ 2.27} & \underline{23.90 $\pm$ 0.05} & \textbf{30.93 $\pm$ 0.02} \\
      \textbf{\proj-node} & \underline{89.52 $\pm$ 0.06} & \underline{62.93 $\pm$ 0.56} & 23.46 $\pm$ 0.08 & \underline{29.69 $\pm$ 0.02}\\
    \hline
    \end{tabular}
    }
    \captionof{table}{\small Performance of node classification (mean in percentage $\pm$ 95$\%$ confidence level). \textbf{Bold font} and {underline} highlight the best performance and the second best performance on average.} 
    
    \label{tab:auc node results}
    \end{minipage}
    \vspace{-7mm}
\end{figure}

\begin{table*}[t]
\vspace{-2.5mm}
\centering
    \resizebox{1.0\textwidth}{!}{%
    \begin{tabular}{r|cccccc|}
    \multicolumn{1}{r}{Method} &   & \multicolumn{1}{c}{Train} & \multicolumn{1}{c}{\textbf{Test}} & \multicolumn{1}{c}{\textbf{Inf.~Lat.}} & \multicolumn{1}{c}{\textbf{CPU (MJ)}} & \multicolumn{1}{c}{GPU (MJ)} \\
    \hline
    
    \cellcolor[gray]{0.95}\cellcolor[gray]{0.95}\cellcolor[gray]{0.95}TGN-trunc & 
 \multirow{10}{*}{\rotatebox[origin=c]{90}
    {Wikipedia}} &\cellcolor[gray]{0.95}\cellcolor[gray]{0.95}\cellcolor[gray]{0.95}13.06 & \cellcolor[gray]{0.95}\cellcolor[gray]{0.95}\cellcolor[gray]{0.95}1.65 & \cellcolor[gray]{0.95}\cellcolor[gray]{0.95}\cellcolor[gray]{0.95}1.20 & \cellcolor[gray]{0.95}\cellcolor[gray]{0.95}\cellcolor[gray]{0.95}2,237 & \cellcolor[gray]{0.95}\cellcolor[gray]{0.95}\cellcolor[gray]{0.95}1.04  \\ 
     TGN-unif &  & 13.25 & 1.67 & 1.23 & 2,194 & 1.02  \\
      \cellcolor[gray]{0.95}\cellcolor[gray]{0.95}\cellcolor[gray]{0.95}TGAT-trunc & & \cellcolor[gray]{0.95}\cellcolor[gray]{0.95}\cellcolor[gray]{0.95}10.68 & \cellcolor[gray]{0.95}\cellcolor[gray]{0.95}\cellcolor[gray]{0.95}1.35 & \cellcolor[gray]{0.95}\cellcolor[gray]{0.95}\cellcolor[gray]{0.95}0.87 & \cellcolor[gray]{0.95}\cellcolor[gray]{0.95}\cellcolor[gray]{0.95}1,882 &\cellcolor[gray]{0.95}\cellcolor[gray]{0.95}\cellcolor[gray]{0.95}0.82\\
      TGAT-unif & & 10.70 &  1.34 & 0.90 & 1,779 & 0.76 \\
      \cellcolor[gray]{0.95}\cellcolor[gray]{0.95}\cellcolor[gray]{0.95}APAN-trunc & &\cellcolor[gray]{0.95}\cellcolor[gray]{0.95}\cellcolor[gray]{0.95}10.67 &\cellcolor[gray]{0.95}\cellcolor[gray]{0.95}\cellcolor[gray]{0.95}1.48 &\cellcolor[gray]{0.95}\cellcolor[gray]{0.95}\cellcolor[gray]{0.95}0.57   & \cellcolor[gray]{0.95}\cellcolor[gray]{0.95}\cellcolor[gray]{0.95}2,034 & \cellcolor[gray]{0.95}\cellcolor[gray]{0.95}\cellcolor[gray]{0.95}0.90  \\
      APAN-unif & & 10.67 & 1.51 & 0.58 & 1,853 & 0.84  \\
      \cellcolor[gray]{0.95}\cellcolor[gray]{0.95}\cellcolor[gray]{0.95}NAT & &\cellcolor[gray]{0.95}\cellcolor[gray]{0.95}\cellcolor[gray]{0.95}14.81 &\cellcolor[gray]{0.95}\cellcolor[gray]{0.95}\cellcolor[gray]{0.95}1.72 & \cellcolor[gray]{0.95}\cellcolor[gray]{0.95}\cellcolor[gray]{0.95}1.11 & \cellcolor[gray]{0.95}\cellcolor[gray]{0.95}\cellcolor[gray]{0.95}2,900 & \cellcolor[gray]{0.95}\cellcolor[gray]{0.95}\cellcolor[gray]{0.95}1.29 \\
     NAT-node & & 12.28 & 1.32 & 0.82 & 2,621 &  1.02 \\
   \cellcolor[gray]{0.95}\textbf{\proj-edge} & &\cellcolor[gray]{0.95}{8.28} & \cellcolor[gray]{0.95}{0.86} & \cellcolor[gray]{0.95}{0.43}  & \cellcolor[gray]{0.95}1,609 & \cellcolor[gray]{0.95}0.85 \\
   \textbf{\proj-node} & &{7.95} & {0.86} & {0.43}  & 1,547 & 0.80 \\ 
    \hline
    \multicolumn{1}{r}{Method} &  &  \multicolumn{1}{c}{Train} & \multicolumn{1}{c}{\textbf{Test}} & \multicolumn{1}{c}{\textbf{Inf.~Lat.}} & \multicolumn{1}{c}{\textbf{CPU (MJ)}} & \multicolumn{1}{c}{GPU (MJ)} \\
    \hline
    
    \cellcolor[gray]{0.95}\cellcolor[gray]{0.95}TGN-trunc & {\multirow{10}{*}{\rotatebox[origin=c]{90}
    {Reddit}}} & \cellcolor[gray]{0.95}\cellcolor[gray]{0.95}51.24 & \cellcolor[gray]{0.95}\cellcolor[gray]{0.95}7.25 & \cellcolor[gray]{0.95}\cellcolor[gray]{0.95}4.74 & \cellcolor[gray]{0.95}\cellcolor[gray]{0.95}9,218 & \cellcolor[gray]{0.95}\cellcolor[gray]{0.95}4.45  \\ 
    TGN-unif &  &  51.97 & 7.38 & 4.82 & 8,853 & 4.64 \\
     \cellcolor[gray]{0.95}\cellcolor[gray]{0.95}TGAT-trunc  & &  \cellcolor[gray]{0.95}\cellcolor[gray]{0.95}38.64 & \cellcolor[gray]{0.95}\cellcolor[gray]{0.95}5.46 & \cellcolor[gray]{0.95}\cellcolor[gray]{0.95}3.86 & \cellcolor[gray]{0.95}\cellcolor[gray]{0.95}7,108 & \cellcolor[gray]{0.95}\cellcolor[gray]{0.95}3.24  \\
     TGAT-unif  & &  40.40 &  5.64 & 4.00 & 7,221 & 3.25  \\
     \cellcolor[gray]{0.95}\cellcolor[gray]{0.95}APAN-trunc  & &  \cellcolor[gray]{0.95}\cellcolor[gray]{0.95}41.32 & \cellcolor[gray]{0.95}\cellcolor[gray]{0.95}6.42 & \cellcolor[gray]{0.95}\cellcolor[gray]{0.95}2.44 & \cellcolor[gray]{0.95}\cellcolor[gray]{0.95}7,732 & \cellcolor[gray]{0.95}\cellcolor[gray]{0.95}3.43  \\
      APAN-unif & & 41.75 & 6.55 & 2.48 & 7,779 & 3.47 \\
      \cellcolor[gray]{0.95}\cellcolor[gray]{0.95}NAT & & \cellcolor[gray]{0.95}\cellcolor[gray]{0.95}54.08 & \cellcolor[gray]{0.95}\cellcolor[gray]{0.95}7.43 & \cellcolor[gray]{0.95}\cellcolor[gray]{0.95}4.82 & \cellcolor[gray]{0.95}\cellcolor[gray]{0.95}11,517 & \cellcolor[gray]{0.95}\cellcolor[gray]{0.95}5.59 \\
     NAT-node & & 46.85 & 6.02 & 3.63 & 10,686 & 4.97  \\
   \cellcolor[gray]{0.95}\textbf{\proj-edge} & &\cellcolor[gray]{0.95}{38.55} & \cellcolor[gray]{0.95}{3.68} & \cellcolor[gray]{0.95}{1.88}  & \cellcolor[gray]{0.95}6,474 & \cellcolor[gray]{0.95}3.86  \\
   \textbf{\proj-node} & &{34.52} & {3.68} & {1.84}  & 6,443 & 3.70 \\ 
    \hline
    \end{tabular}
    \begin{tabular}{cccccc|}
     & {Train} & \multicolumn{1}{c}{\textbf{Test}} & \multicolumn{1}{c}{\textbf{Inf.~Lat.}} & \multicolumn{1}{c}{\textbf{CPU (MJ)}} & \multicolumn{1}{c}{GPU (MJ)} \\
    \hline
    
    \multirow{10}{*}{\rotatebox[origin=c]{90}
    {Ubuntu}}
      & \cellcolor[gray]{0.95}15.95 & \cellcolor[gray]{0.95}2.05 & \cellcolor[gray]{0.95}1.37 & \cellcolor[gray]{0.95}3,332 & \cellcolor[gray]{0.95}1.75   \\
     &  15.55 & 2.11 & 1.42 & 3,326 & 1.78  \\
     & \cellcolor[gray]{0.95}12.37 & \cellcolor[gray]{0.95}1.70 & \cellcolor[gray]{0.95}1.20 & \cellcolor[gray]{0.95}3,028 &\cellcolor[gray]{0.95}1.30 \\
     &  13.64 & 1.93 & 1.27 & 3,254 & 1.37  \\
     & \cellcolor[gray]{0.95}12.96 & \cellcolor[gray]{0.95}1.94 & \cellcolor[gray]{0.95}0.61 & \cellcolor[gray]{0.95}2,802 &\cellcolor[gray]{0.95}1.36 \\
     &  13.01 & 2.07 & 0.62 & 2,894 & 1.34 \\
     &  \cellcolor[gray]{0.95}17.52 & \cellcolor[gray]{0.95}1.96 & \cellcolor[gray]{0.95}1.22 & \cellcolor[gray]{0.95}2,853 & \cellcolor[gray]{0.95}1.27\\
     &  15.60 & 1.60 & 1.30 & 2,824 & 1.23 \\
  & \cellcolor[gray]{0.95}{11.06} & \cellcolor[gray]{0.95}{1.30} & \cellcolor[gray]{0.95}{0.57}  & \cellcolor[gray]{0.95}2,764 & \cellcolor[gray]{0.95}2.24  \\ 
   & {10.74} & {1.35} & {0.59}  & 2,748 & 2.17 \\ 
    \hline
      & {Train} & \multicolumn{1}{c}{\textbf{Test}} & \multicolumn{1}{c}{\textbf{Inf.~Lat.}} & \multicolumn{1}{c}{\textbf{CPU (MJ)}} & \multicolumn{1}{c}{GPU (MJ)} \\
    \hline
    \multirow{10}{*}{\rotatebox[origin=c]{90}
    {Wiki-talk}}
      & \cellcolor[gray]{0.95}74.63 & \cellcolor[gray]{0.95}11.76 & \cellcolor[gray]{0.95}7.61 & \cellcolor[gray]{0.95}16,715 & \cellcolor[gray]{0.95}9.57\\
     & 85.96 & 13.77 & 8.19 & 18,925 & 11.2\\
      & \cellcolor[gray]{0.95}58.37 & \cellcolor[gray]{0.95}8.74 & \cellcolor[gray]{0.95}5.88 & \cellcolor[gray]{0.95}14,787 & \cellcolor[gray]{0.95}5.90\\
      & 63.32 & 9.14 & 6.35 & 16,553 & 6.75 \\
      & \cellcolor[gray]{0.95}60.68 & \cellcolor[gray]{0.95}10.34 & \cellcolor[gray]{0.95}2.98 & \cellcolor[gray]{0.95}13,898 & \cellcolor[gray]{0.95}6.94  \\
      & 64.06 & 10.61 & 2.88 & 14,463 & 7.04 \\
      & \cellcolor[gray]{0.95}81.32 & \cellcolor[gray]{0.95}10.03 & \cellcolor[gray]{0.95}6.08 & \cellcolor[gray]{0.95}14,736 & \cellcolor[gray]{0.95}7.13\\
      & 67.28 & 8.79 & 4.88 & 14,492 & 6.82\\
   &\cellcolor[gray]{0.95}{59.18} & \cellcolor[gray]{0.95}{7.30} & \cellcolor[gray]{0.95}{3.03} & \cellcolor[gray]{0.95}12,277 & \cellcolor[gray]{0.95}15.0 \\
   &  {56.36} & {7.01} & {2.99}   & 11,519 & 13.6 \\
    \hline
    \end{tabular}

    \begin{tabular}{cccccc}
      & \multicolumn{1}{c}{Train} & \multicolumn{1}{c}{\textbf{Test}} & \multicolumn{1}{c}{\textbf{Inf.~Lat.}} & \multicolumn{1}{c}{\textbf{CPU (MJ)}} & \multicolumn{1}{c}{GPU (MJ)} \\
    \hline
    \multirow{10}{*}{\rotatebox[origin=c]{90}
    {GDELT}}
      &\cellcolor[gray]{0.95}\cellcolor[gray]{0.95}\cellcolor[gray]{0.95}1977 &\cellcolor[gray]{0.95}406 &\cellcolor[gray]{0.95}302 &\cellcolor[gray]{0.95}431,567 & \cellcolor[gray]{0.95}174.6 \\
     & 4791 & 1013 & 777 & 797,311 & 389.4 \\
      & \cellcolor[gray]{0.95}1123 &\cellcolor[gray]{0.95}248 &\cellcolor[gray]{0.95}183  &\cellcolor[gray]{0.95}306,860 &\cellcolor[gray]{0.95}92.7  \\
       & 1493 & 383 & 301 &  251,452 & 83.8  \\
       &\cellcolor[gray]{0.95}1530 &\cellcolor[gray]{0.95}338 &\cellcolor[gray]{0.95}200 & \cellcolor[gray]{0.95}485,818 &\cellcolor[gray]{0.95}211.2  \\
       & 1839 & 347 & 187 & 512,061 & 188.4 \\
      &\cellcolor[gray]{0.95}- &\cellcolor[gray]{0.95}- &\cellcolor[gray]{0.95}- &\cellcolor[gray]{0.95}- &\cellcolor[gray]{0.95}- \\
      & - & - & - & - & - \\
  & \cellcolor[gray]{0.95}1323 & \cellcolor[gray]{0.95}139 & \cellcolor[gray]{0.95}63  &  \cellcolor[gray]{0.95}252,267 & \cellcolor[gray]{0.95}307.6 \\
  & {1089} & {118} &{60}  &215,820 &258.5 \\
    \hline
      & \multicolumn{1}{c}{Train} & \multicolumn{1}{c}{\textbf{Test}} & \multicolumn{1}{c}{\textbf{Inf.~Lat.}} & \multicolumn{1}{c}{\textbf{CPU (GJ)}} & \multicolumn{1}{c}{GPU (GJ)} \\
    \hline
    \multirow{10}{*}{\rotatebox[origin=c]{90}
    {MAG}}
       & \cellcolor[gray]{0.95}6845 &\cellcolor[gray]{0.95}1266 &\cellcolor[gray]{0.95}795 &\cellcolor[gray]{0.95}15,041 &\cellcolor[gray]{0.95}1.59 \\
     & 5704 & 1224 & 741 & 21,334 & 2.02 \\
       
       &\cellcolor[gray]{0.95}3465 &\cellcolor[gray]{0.95}754 &\cellcolor[gray]{0.95}518 &\cellcolor[gray]{0.95}10,427 &\cellcolor[gray]{0.95}1.06 \\
       & 3520 & 774 & 541 & 13,236 & 1.35  \\
      & \cellcolor[gray]{0.95}12347 & \cellcolor[gray]{0.95}1600 & \cellcolor[gray]{0.95}809 &\cellcolor[gray]{0.95}22,262 &\cellcolor[gray]{0.95}2.45  \\
      & - & - & - & - & - \\
      &\cellcolor[gray]{0.95}- &\cellcolor[gray]{0.95}- &\cellcolor[gray]{0.95}- & \cellcolor[gray]{0.95}- & \cellcolor[gray]{0.95}- \\
      & - & - & - & - & - \\
   &   \cellcolor[gray]{0.95}3073 & 
 \cellcolor[gray]{0.95}593 & \cellcolor[gray]{0.95}296 & \cellcolor[gray]{0.95}2,684 & \cellcolor[gray]{0.95}1.73  \\
   & 2729 & 577 & 279  &2,870 & 1.69 \\
    \hline
    \end{tabular}
    }
    \vspace{-2mm}
    \caption{\small Scalability evaluation on all datasets. Note that MJ = $10^6$ Joules, and GJ = $10^9$ Joules. Note that TGL is adopted to implement baselines including TGN, TGAT, and APAN.}
    \vspace{-8mm}
    \label{tab:scalability}
\end{table*}
\vspace{-3.0mm}
\subsection{Results and Discussion}
 \vspace{-2.0mm}
\textbf{Prediction performance.} 
The node classification performance is given in Table~\ref{tab:auc node results} and the link prediction performance in AUC is given in Table~\ref{tab:auc results}. The link prediction results in AP and in MRR are given in Appendix Table~\ref{apx:ap} and Table~\ref{apx:mrr} respectively. 

For link prediction, \proj-edge achieves either the best or the second best performances in 10 out of 12 settings (6 datasets with inductive and transductive tasks), while \proj-node closely follows or sometimes marginally surpasses \proj-edge. NAT is the best performing baseline for link prediction because it leverages the joint neighborhood structural features~\cite{li2020distance,wang2020inductive} designed for node-pairs. However, NAT cannot be applied to node classification, and it consumes significant GPU memory. We discuss NAT-node in detail later in this section. 
TGN-trunc and TGN-unif are the second most competitive baselines. They achieve similar performance to \proj-edge and \proj-node sometimes but significantly fall behind on the datasets Ubuntu and Wiki-talk, and 3 of the 4 node classification tasks. 
One aspect that differentiate Ubuntu and Wiki-talk to the rest of the dataset is that they do not have node or edge features (Table~\ref{tab:dataset}). We hypothesize that the benefit of recent sampling is more obvious with the absence of additional features because \proj can gather effective historical information from  diverse and recent temporal neighbors. The other baselines usually perform far behind \proj.


For node classification, both \proj-edge and \proj-node achieve similar or better results than the baselines. In MAG, all baselines fail to perform well and we cannot reproduce the scores reported in~\citet{zhou2022tgl} (more discussion in Appendix~\ref{apx:tgl_mag}). 

\proj-edge often outperforms \proj-node because \proj-edge tends to encode more temporal neighbors (node IDs plus timestamps) while \proj-node only encodes neighbors with unique node IDs.

\textbf{Time and energy efficiency.} 
The details of the scalability metrics are shown in Table~\ref{tab:scalability}. 

\proj-node always have faster training, testing and lower energy consumption than all baselines while \proj-edge closely follows. \proj-node trains 1.63-2.04$\times$ faster than NAT and 1.32-4.40$\times$ faster than TGN-unif and TGN-trunc. 

On inference latency, except for APAN on the Wiki-talk dataset, both \proj-node and \proj-edge shows significant improvements over all settings. APAN aims to improve inference latency but consumes massive CPU memory. On large-scale datasets such as GDELT and MAG, APAN requires significant latency for transfering messages from CPU to GPU and no longer has advantages. 
Apart from APAN, \proj-node gives 1.63-12.95$\times$ speedup on inference latency over all datasets.



Although \proj's GPU energy consumption is slightly larger, it is negligible compared to the CPU energy consumption. Compared to \proj-edge, NAT consumes up to 1.8$\times$ more energy while TGN-trunc and TGN-unif consumes 1.2-5.6$\times$ and  1.2-7.94$\times$ more energy over all datasets respectively. On billion-scale datasets MAG, \proj-edge is 3.88-8.29$\times$ more energy efficient than all methods. 
The large difference between the energy consumed by the CPU v.s.~the GPU can be attributed to (1) the default energy costs by CPU when no application is running, and (2) the intensive CPU-GPU communications which is counted toward the CPU cost as it controls the communication (more details in Appendix \ref{apd:energy}).

Over all baselines, truncation is usually more time and energy efficient than uniform sampling. The advantage can be seen more obviously in inference latency. However, since APAN can perform the sampling asynchronously, its inference latency is similar across these two sampling strategies.

    
We further compare the efficiency of different sampling approaches in Appendix~\ref{apd:sampling}, which shows that forward sampling is significantly faster. We also plot the model convergence curves in Fig.~\ref{fig:converge} on the link prediction validation performance v.s.~CPU/GPU
wall-clock training time on Reddit. \proj-edge and \proj-node are among the fastest to converge to a high performance.

\textbf{Comparisons between \proj and NAT-node.} Overall, NAT-node shows improvement over NAT on all of the scalability metrics. However, it does not outperform \proj in either computational efficiency or prediction performance. As shown by Table~\ref{tab:scalability}, over all datasets, NAT-node is still slower than \proj-edge (1.14-1.48$\times$ slower in training and 1.61-1.91$\times$ slower in inference). It still causes GPU out-of-memory error for large-scale datasets (GDELT and MAG). We think the computational inefficiency may be caused by the storage and maintenance of their neighborhood representations which are primarily used for structural features construction. Furthermore, NAT-node does not perform as well as \proj-edge on link prediction (Table~\ref{tab:auc results}) and node classification (Table~\ref{tab:auc node results}).
 
 \vspace{-2mm}
\subsection{Additional analysis}

\textbf{The effect of different $\alpha$'s 
 and the optimality of recent sampling.} We study how changes in the replacement probability $\alpha$ affects the link prediction performance. We show the transductive task on Reddit in Fig.~\ref{fig:alpha} Left (with inductive task in Appendix Fig.~\ref{fig:alpha_reddit_inductive}) and the inductive task on Ubuntu in Fig.~\ref{fig:alpha} Right (with transductive task in Appendix Fig.~\ref{fig:alpha_ubuntu_transductive}). We observe a general trend for both \proj-edge and \proj-node that increasing $\alpha$ from 0.2 to 0.8 or 0.9 improves the performance. However, when $\alpha=1$, the performance drops significantly. When $\alpha=1$, new interactions will always replace the existing down-sampled temporal neighbors, which is similar to the strategy of truncation. 
 This demonstrates the sub-optimality of truncation and the necessity of randomness in neighbor sampling. For smaller $\alpha$, less recent neighboring nodes will be kept in the hash table, which performs more like uniform sampling. The performance decay for small $\alpha$ indicates the sub-optimality of uniform sampling. Therefore, recent sampling is important for prediction performance.

\textbf{The effect of the hash table size $s$.} In Table~\ref{tab:s_nlb}, we show how changes in $s$ affect the performance and scalability of \proj-edge and \proj-node respectively using the Reddit Dataset. While $s=0$ is usually the worst-performing in both transductive and inductive link prediction, it has similar performance in node classification and it is more time and energy-efficient as expected. In contrast, when $s=5,10$, the performance on node classification achieves the worst for both \proj-edge and \proj-node. Then, both performance and scalability metrics have an increasing trend as $s$ gets larger. However, the performance for \proj-node is still increasing at $s=30$, while \proj-edge shows decreases. The link prediction performance of \proj-edge is better than \proj-node but the node classification performance of \proj-node is better. We hypothesize that \proj-edge can more easily overfit data as $s$ becomes large. Finally, \proj-node is always slightly more time and energy-efficient than \proj-edge.

We give the evaluation of MRR for various $\alpha$'s and $s$' in Appendix Table~\ref{MRR_alpha} and Table~\ref{MRR_s} respectively.

\begin{figure}
\begin{minipage}{0.48\textwidth}
\centering 
\includegraphics[trim={0.4cm 0.3cm 0.4cm 0.3cm}, clip, width=\textwidth]{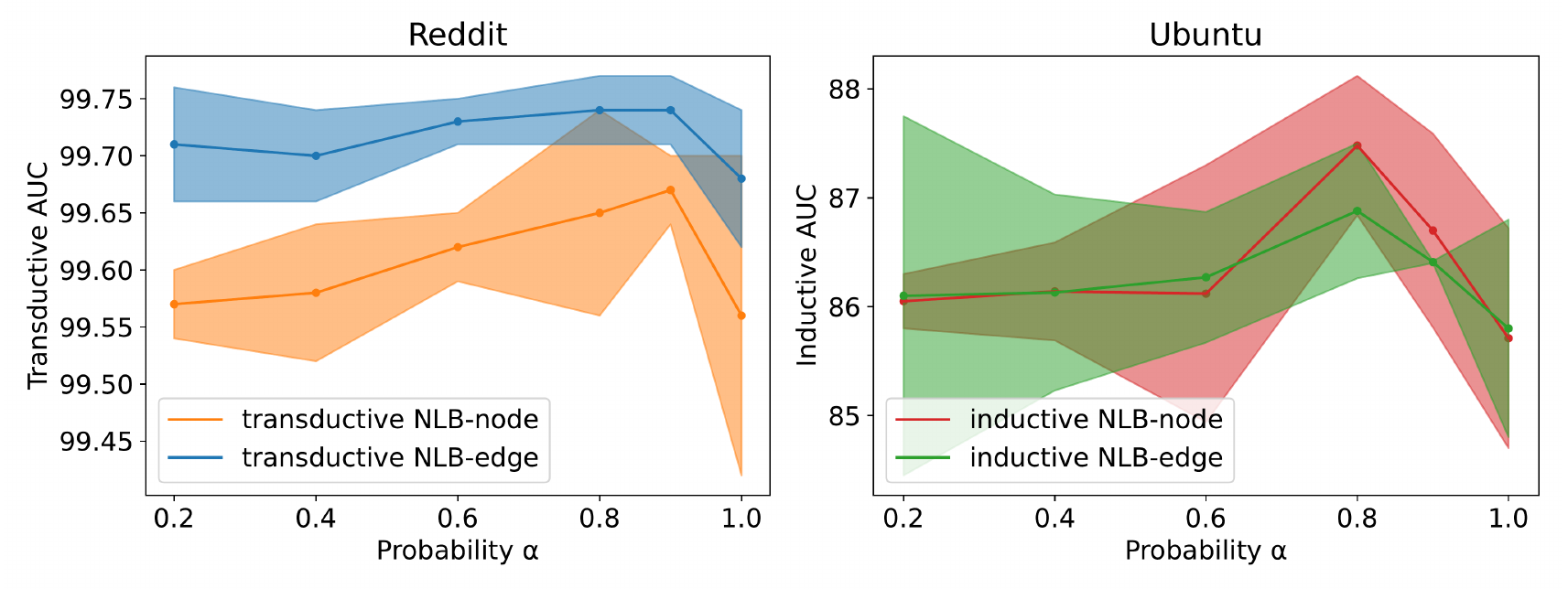}
    \vspace{-0.3cm}
    \caption{\small{The changes in transductive link prediction test performance on Reddit (Left) and in inductive link prediction test performance on Ubuntu (Right) with respect to $\alpha$'s.}}
    \label{fig:alpha}
\end{minipage}
\hfill
\begin{minipage}{0.49\textwidth}
    \centering
    
    \resizebox{\textwidth}{!}{%
    \begin{tabular}{c|c|cccccccc}
    \hline
      \multicolumn{1}{c|}{}&\multicolumn{1}{c|}{s} & \multicolumn{1}{c}{Transductive} & \multicolumn{1}{c}{Inductive} &\multicolumn{1}{c}{Node Class.} &
     \multicolumn{1}{c}{Train} & \multicolumn{1}{c}{Test} & \multicolumn{1}{c}{Inf.~Lat.} & \multicolumn{1}{c}{CPU}& \multicolumn{1}{c}{GPU} \\
    \hline
      \multirow{7}{*}{\rotatebox[origin=c]{90}
    {NLB-edge}} & 0 & 99.59 $\pm$ 0.11 & 98.42 $\pm$ 0.17 & 62.15 $\pm$ 0.49 & 34.26 & 4.13 & 1.87 & 5,699 & 2.88\\
     \multicolumn{1}{c|}{}&5 & 99.68 $\pm$ 0.03 & 99.24 $\pm$ 0.04 & 60.53 $\pm$ 0.53 & 39.97 & 4.90 & 2.22 & 6,600 & 3.57 \\
      \multicolumn{1}{c|}{}& 10 &99.69 $\pm$ 0.04 & 99.44 $\pm$ 0.13 & 60.33 $\pm$ 1.32 & 40.08 & 4.87 & 2.21 & 6,600 & 3.82\\
     \multicolumn{1}{c|}{}&15 &  99.65 $\pm$ 0.10  & 99.32 $\pm$ 0.04 & 62.34 $\pm$ 1.94  & 39.49 & 4.86 & 2.20  & 6,576 & 4.01 \\
     \multicolumn{1}{c|}{}&\underline{20} & 99.74 $\pm$ 0.03 & 99.43 $\pm$ 0.07 & 62.73 $\pm$ 2.27  & 39.10 & 4.87 & 2.19 & 6,678  & 4.10 \\
     \multicolumn{1}{c|}{}&25 & 99.72 $\pm$ 0.07 & 99.47 $\pm$ 0.21 & 60.87 $\pm$ 0.14 & 39.67 & 4.94 & 2.24 & 6,666 & 4.42 \\
    \multicolumn{1}{c|}{}& 30 & 99.67 $\pm$ 0.04 & 99.25 $\pm$ 0.09 & 61.30 $\pm$ 0.95 & 41.32 & 4.89 & 2.23 & 6,930 & 4.68 \\
    \hline
    \multirow{7}{*}{\rotatebox[origin=c]{90}
    {NLB-node}}& 0 & 99.59 $\pm$ 0.11 & 98.42 $\pm$ 0.17 & 62.15 $\pm$ 0.49 & 34.26 & 4.13 & 1.87 & 5,699 & 2.88 \\
    \multicolumn{1}{c|}{}& 5 & 99.60 $\pm$ 0.04 & 99.20 $\pm$ 0.02 & 61.64 $\pm$ 4.03 & 39.03 & 4.87  & 2.18 & 6,460 & 3.30\\
     \multicolumn{1}{c|}{}& 10 & 99.59 $\pm$ 0.04 & 99.16 $\pm$ 0.17 & 59.16 $\pm$ 2.27 & 39.09 & 4.85 & 2.18 & 6,432 & 3.56\\
    \multicolumn{1}{c|}{}& 15 & 99.66 $\pm$ 0.07 & 99.23 $\pm$ 0.08  & 61.96 $\pm$ 3.15 & 39.32 & 4.86 & 2.22 & 6,447 & 3.72\\
    \multicolumn{1}{c|}{}& \underline{20} & 99.67 $\pm$ 0.02 & 99.36 $\pm$ 0.09 & 62.93 $\pm$ 0.56 & 39.09 & 4.86 & 2.18 & 6,541 & 3.79\\
     \multicolumn{1}{c|}{}&25 & 99.66 $\pm$ 0.07 & 99.37 $\pm$ 0.07 & 62.30 $\pm$ 1.11 & 39.33 & 4.91 & 2.21 & 6,563 & 3.88\\
     \multicolumn{1}{c|}{}&30 & 99.65 $\pm$ 0.01 & 99.38 $\pm$ 0.05 & 66.30 $\pm$ 1.62 & 40.07 & 4.88 & 2.19 & 6,739 & 4.08\\
    \hline
    \end{tabular}
    }
    \captionof{table}{ \small The performance and scalability of \proj-edge and \proj-node on the Reddit dataset given different down-sampled neighbor hash table sizes $s$. $s=20$ is the setting used for comparison with baselines. CPU and GPU energies are measured in MJ.}
    \label{tab:s_nlb}
    \end{minipage}
    \vspace{-5mm}
\end{figure}

 \vspace{-3mm}
\section{Limitations and Future Work}
 \vspace{-2mm}
\label{sec:limit} 
We are aware of the limitation of \proj and highlight some future research directions. For industry-level graphs, it may not be possible to maintain the down-sampled temporal neighbors of all nodes in a single GPU. 
In the future, we plan to adopt distributed GPUs by partitioning the down-sampled neighbors of different nodes into different GPUs. We will guarantee that the memory used for each node is contained within the same GPU so that the maintenance can still be fast.

 \vspace{-3mm}
\section{Conclusion}
\vspace{-2mm}
\label{sec:conclusion}
In this work, we introduced \proj which abandons the traditional time-consuming temporal neighbor backtracking and sampling while adopting the newly proposed forward recent sampling. 
We proved that \proj-edge achieves recent sampling of links and \proj-node achieves recent sampling of nodes. Our extensive
experiments demonstrate that \proj is effective in both prediction performance and scalability while significantly improving inference latency and energy efficiency.

\section{Acknowledgement}
The authors would like to thank Hongkuan Zhou of USC for the time to discuss the code for TGL~\citep{zhou2022tgl}. We would also like to thank all reviewers for providing valuable feedback for improving this work.
Y.~Luo and P.~Li are supported by NSF awards PHY-2117997, IIS-2239565.
\bibliographystyle{unsrtnat}
\bibliography{log_2024}

\appendix
\newpage
\appendix

\section{The proof of Theorem~\ref{thm:edge} (\proj-edge achieves recent sampling)}\label{apx:nlbedgeproof}

\begin{theorem}[\proj-edge achieves recent sampling (Theorem~\ref{thm:edge} Restated)] 
Suppose links come in for any node (e.g.~$u$) by following a Poisson point process with a constant intensity (e.g.~$\lambda$), and suppose a temporal neighbor $(v, e_{u,v}^{t_i},t_i)$ is inserted into $S_u$ at time $t_i$, then $\Pr((v,e_{u,v}^{t_i},t_i) \in S_u^t) = \exp(\frac{\alpha\lambda}{s}(t_i - t))$.
\end{theorem}

\textit{Proof.}
Let $N_i \coloneqq (v_i, e^{t_i}_{u,v_i}, t_i)$ and $N_i \in \mathcal{N}^t_u$  be one of the historical temporal neighbors of $u$, where $i \in \bigg[1, |\mathcal{N}^t_u|\bigg]$. By construction, $N_i$ is hashed to the position $\emph{hash}(v_i, t_i)$ (Eq.~\ref{eq:hash_e}).
The probability that any other temporal neighbors is inserted to $S_u$ is $\alpha$ as we define in Sec.~\ref{sec:sampling}.
Since we suppose $N_i$ is already inserted to $S_u$, we only need to evaluate the probability that it does not get replaced by another temporal neighbor. We know from the Poisson process that \begin{align}
    \mathbb{E}\bigg[\text{\# links of $u$ arrive since } t_i\bigg] = \lambda (t-t_i).
\end{align}
On average, each link has probability $\frac{\alpha}{s}$ of replacing $N_i$. 
Then,
\begin{align}
    \mathbb{E}\bigg[\text{\# links replace } S_u[\emph{hash}(v_i, t_i)] \text{ since } t_i\bigg] = \frac{\alpha\lambda(t-t_i)}{s}.
\end{align}
 Thus, the intensity of links being inserted to the same position as $N_i$ is $\frac{\alpha\lambda}{s}$.

By the property of a Poisson process, the probability that none of the links get inserted to the same position as $N_i$ from $t_i$ to $t$ is $\exp (\frac{\alpha\lambda}{s} (t_i - t))$. Overall, 
\begin{align}
    \Pr(N_i \in S_u^t) = 
    \exp(\frac{\alpha\lambda}{S} (t_i - t)),
\end{align}
completing the proof.

We have shown that $\Pr((v,e_{u,v}^{t_i},t_i) \in S_u^t)$ is proportional to $\exp(c(t_i - t))$ for $c \ge 0$. Thus, \proj-edge achieves recent sampling.

\section{The sample properties of \proj-node}~\label{apx:nlb-node}
In this section, we first propose a new sampling strategy called \emph{recent node-wise sampling}, we then show that \proj-node essentially achieves recent node-wise sampling in $O(1)$.

\textbf{Recent node-wise sampling.}  Instead of sampling temporal neighbors, recent node-wise sampling considers sampling unique neighboring nodes: Given a center node $u$ and its neighboring node $v$, only the latest interactions till time $t$ between $u$ and $v$ can be sampled and the down-sampled neighbor nodes do not contain duplicates.
This strategy can speed up the aggregation of a node's neighborhood as it reduces the number of sampled candidates.

We now prove that \proj-node achieves recent node-wise sampling.
\begin{theorem}[\proj-node achieves recent node-wise sampling]\label{thm:node}  

Suppose each unique node (e.g.~$v_i \in \mathcal{V}$) interacts with a center node (e.g.~$u$) by following a different Poisson point process with a constant intensity (e.g.~$\lambda_{v_i}$), and suppose the latest interaction between a center node and a neighboring node, (e.g.~$(v_i, e^{t_i}_{u,v_i}, t_i)$) is inserted into $S_u$ at time $t_i$, then $\Pr((v_i, e^{t_i}_{u,v_i}, t_i) \in S_u^t) = \prod_{v_j\in~ \mathcal{V} \setminus \{v_i\} } \bigg(\frac{s-1}{s} + \frac{1}{s} \exp(\alpha\lambda_{v_j} (t_i - t))\bigg)$.

\end{theorem}
\textit{Proof.}
The probability that any other temporal neighbors is inserted to $S_u$ is $\alpha$ as we define in Sec.~\ref{sec:sampling}.
Similar to the argument in Theorem~\ref{thm:edge}, since we suppose the latest link with between node $v_i$ and $u$, i.e., $(v_i, e^{t_i}_{u,v_i}, t_i)$, is already inserted to $S_u$, we only need to evaluate the probability that it does not get replaced by another node in $\mathcal{V}$.
\proj-node
computes the hash values based on neighbor node IDs without timestamps. Thus, the probability a different node $v_j$ has the same hash value as $v_i$ is the following, \begin{align}
    \Pr (\emph{hash}(v_j) = \emph{hash}(v_i)) = \frac{1}{s}.
\end{align}
We also have \begin{align}
    \Pr (\emph{hash}(v_j) \ne \emph{hash}(v_i)) = \frac{s - 1}{s}.
\end{align}
For an arbitrary $v_j \in \mathcal{V}$ where $\emph{hash}(v_i) = \emph{hash}(v_j)$, during $(t_i, t)$, \begin{align}
    \mathbb{E}\bigg[\text{\# links between $u$ and $v_j$ arrive since } t_i\bigg] = \lambda_{v_j} (t-t_i).
\end{align}
Notice that if $v_j$ never interacts with $u$, then $\lambda_{v_j} = 0$. Suppose there are interactions between $u$ and $v_j$, then each link between $u$ and $v_j$ has probability $\alpha$ to collide with $(v_i, e^{t_i}_{u,v_i}, t_i)$ stored in $S_u^t$. Then,
\begin{align}
    \mathbb{E}\bigg[\text{\# links between $u$ and $v_j$ replace } S_u[\emph{hash}(v_i)] \text{ since } t_i\bigg] = \alpha\lambda_{v_j}(t-t_i).
\end{align}
Thus, the intensity of links between $u$ and $v_j$ being inserted to $S_u[\emph{hash}(v_i)]$ is $\alpha\lambda_{v_j}$. The probability none of the links associated with node $v_j$ is inserted during $(t_i, t)$ is then $\exp(\alpha\lambda_{v_j}(t_i - t))$, according to the property of a Poisson process.

For the event $(v_i, e^{t_i}_{u,v_i}, t_i)\in S_u^t$ to happen, it has to be true that for all node $v_j$ that $v_j \ne v_i$, either $\text{hash}(v_j) \ne \text{hash}(v_i)$  or none of the links associated with node $v_j$ is inserted to $S_u$ during $(t_i, t)$. Overall, 
\begin{align}
    \Pr((v_i, e^{t_i}_{u,v_i}, t_i) \in S_u^t) &= \prod_{v_j\in~ \mathcal{V} \setminus \{v_i\} } \bigg(\frac{s-1}{s} + \frac{1}{s} \exp(\alpha\lambda_{v_j} (t_i - t))\bigg),
\end{align}
completing the proof.

This achieves recent node-wise sampling because as the difference between the current timestamp $t$ and the timestamp $t_i$ of the lastest interaction with a unique neighboring node gets larger, $\Pr((v_i, e^{t_i}_{u,v_i}, t_i) \in S_u^t)$ decays exponentially. The decay factor is controllable by $\alpha$, which is similar to the constant $c$ in recent sampling of edges (Def.~\ref{def:recent}). 
As $\alpha$ approaches 0, the probability for each node of being sampled is approaching a constant regardless of recency, which is similar to uniform sampling. As $\alpha$ approaches 1, as long as there is collision, older nodes in $S_u$ will always be replaced by more recent nodes, which is similar to truncation.

\section{T-encoding}\label{apd:tencoding}
%
Our temporal features are encoded via a T-encoding technique with the definition below.
\begin{definition}[T-encoding]\label{def:tencoding}
For any input timestamps $t_i$, we adopt Fourier features to encode them before using them as features, i.e., with learnable parameter $\omega_i$'s, $1\leq i\leq d$, \begin{align}
    \text{T-encoding}(t) = [\cos(\omega_1 t), \sin(\omega_1 t), ..., \cos(\omega_d t), \sin(\omega_d t)].
\end{align}
\end{definition}
 It has shown to be effective for TGRL~\cite{xu2019self,kazemi2019time2vec,xu2020inductive, wang2020inductive, tgn_icml_grl2020,liu2021neural}.

\section{Dataset Description}
\label{apd:dataset}

\begin{table}
\centering 
\resizebox{0.6\textwidth}{!}{%
\begin{tabular}{l|ccccccc}
\hline
\textbf{Measurement}   & \textbf{Wikipedia} & \textbf{Reddit} & \textbf{GDELT} & \textbf{MAG} & \textbf{Ubuntu} & \textbf{Wiki-talk} \\ \hline
nodes         &  9K & 11K & 17K & 122M & 159K  & 1.1M\\
temporal links   &  157K & 672K & 191M &  1.3B & 964K  & 7.8M\\
node classes &  2 & 2 & 81 & 152 & 0 & 0\\
labels & 217 & 366 & 42M & 1.4M & 0 & 0\\
node features  & 0 &  0 &  413  & 768  & 0  &  0 \\
edge features  & 172 &   172 &   186 & 0 &  0 &  0\\

\hline
\end{tabular}%
}
\caption{\small Summary of dataset statistics.}
\label{tab:dataset}

\end{table}

\begin{table*}[t]
    \centering
    \resizebox{0.7\textwidth}{!}{%
    \begin{tabular}{c|l|cccccc}
    \hline
    \multicolumn{1}{c|}{Task} & Method & \multicolumn{1}{c}{Wikipedia} & \multicolumn{1}{c}{Reddit}& \multicolumn{1}{c}{GDELT}  & \multicolumn{1}{c}{MAG} & \multicolumn{1}{c}{Ubuntu} & \multicolumn{1}{c}{Wiki-talk}\\
    \hline
    \multirow{10}{*}{\rotatebox[origin=c]{90}
    {Transductive}}
    & TGN-trunc & \underline{99.44 $\pm$ 0.08} & {99.59 $\pm$ 0.09} & \underline{98.34 $\pm$ 0.10} & \underline{99.30 $\pm$ 0.06}  & 81.61 $\pm$ 0.20 & {87.03 $\pm$ 2.47}\\
     & TGN-unif & \underline{99.36 $\pm$ 0.04} & {99.60 $\pm$ 0.05} & 96.96 $\pm$ 0.05 & {99.21 $\pm$ 0.04}  & {83.27 $\pm$ 0.96} & {87.24 $\pm$ 0.16}\\
      & TGAT-trunc & {97.65 $\pm$ 0.04} & {97.98 $\pm$ 0.03} & {98.12 $\pm$ 0.01} & {99.08 $\pm$ 0.03}  & 80.62 $\pm$ 0.39 & 84.81 $\pm$ 0.06 \\
       & TGAT-unif & {94.61 $\pm$ 0.23} & {98.12 $\pm$ 0.15} & 97.54 $\pm$ 0.01 & {99.01 $\pm$ 0.05}  & 81.82 $\pm$ 0.12 & 84.94 $\pm$ 0.01\\
        & APAN-trunc & {99.04 $\pm$ 0.03} & {95.73 $\pm$ 2.96} & 96.79 $\pm$ 0.32  & {87.85 $\pm$ 1.68}  & 72.20 $\pm$ 3.94 & {84.13 $\pm$ 4.44}\\
         & APAN-unif & {97.14 $\pm$ 0.67} & {96.57 $\pm$ 0.24} & 96.21 $\pm$ 1.35 & {CPU OOM}  & 78.61 $\pm$ 1.61 & {85.68 $\pm$ 6.00}\\
           & NAT & \textbf{99.72 $\pm$ 0.02} & \textbf{99.89 $\pm$ 0.01} & GPU OOM & GPU OOM  &  \textbf{90.88 $\pm$ 0.19} &  \textbf{95.18 $\pm$ 0.03}\\
           & NAT-node & {99.13 $\pm$ 0.12} & \underline{99.72 $\pm$ 0.02} &  GPU OOM  & GPU OOM  & \underline{87.75 $\pm$ 0.55} & \underline{93.34 $\pm$ 0.55}\\
     \multicolumn{1}{c|}{} & \textbf{\proj-edge} & {99.30 $\pm$ 0.11} & {99.69 $\pm$ 0.02} & \textbf{98.79 $\pm$ 0.39} & \textbf{99.38  $\pm$ 0.02} & \underline{87.52 $\pm$ 0.48}  & {92.05  $\pm$ 0.62} \\
     \multicolumn{1}{c|}{} & \textbf{\proj-node} & {98.77 $\pm$ 0.09} & {99.60 $\pm$ 0.04} & \textbf{98.62  $\pm$  0.19} & { 99.10 $\pm$ 0.02} & \underline{87.61 $\pm$ 0.69}  & {92.11 $\pm$ 0.08} \\
    \hline
    \multicolumn{1}{c|}{\multirow{10}{*}{\rotatebox[origin=c]{90}
    {Inductive}}}
    & TGN-trunc &  \underline{98.42 $\pm$ 0.08} & \underline{99.38 $\pm$ 0.07} & \textbf{98.38 $\pm$ 0.04} & {96.99 $\pm$ 0.08}  &  
  82.27 $\pm$ 1.17 & {88.46 $\pm$ 0.64}\\
     & TGN-unif &  {97.99 $\pm$ 0.13} & {99.21 $\pm$ 0.03} & \textbf{98.48 $\pm$ 0.15} & {97.21 $\pm$ 0.14}  &  \textbf{85.70 $\pm$ 1.58}  & 88.10 $\pm$ 0.13\\
  & TGAT-trunc &  {96.75 $\pm$ 0.02} & {95.77 $\pm$ 0.21} & 94.64 $\pm$ 0.01 & \underline{98.78 $\pm$ 0.01}  & 81.53 $\pm$ 0.14 &  84.10 $\pm$ 0.05\\
   & TGAT-unif &  {92.94 $\pm$ 0.15} & {96.33 $\pm$ 0.09} & 93.32 $\pm$ 0.02  & {98.75 $\pm$ 0.01}  & 82.17 $\pm$ 0.08 & 82.21 $\pm$ 0.05\\
    & APAN-trunc &  {95.18 $\pm$ 0.91} & {96.02 $\pm$ 2.06} & \underline{97.99 $\pm$ 0.03} & {CPU OOM}  &  68.60 $\pm$ 2.41  & 80.47 $\pm$ 0.35\\
     & APAN-unif &  {95.73 $\pm$ 0.77} & {96.03 $\pm$ 1.31} & 96.55 $\pm$ 0.31 & CPU OOM  &  78.64 $\pm$ 2.66 & {88.82 $\pm$ 0.28}\\
       & NAT & \textbf{99.48 $\pm$ 0.02} & \textbf{99.76 $\pm$ 0.03} &  GPU OOM & GPU OOM  & \textbf{86.69 $\pm$ 1.03} &  \textbf{93.27 $\pm$ 0.88}\\
       & NAT-node & {97.78 $\pm$ 0.42} & \underline{99.36 $\pm$ 0.44} &  GPU OOM  & GPU OOM  & 82.68 $\pm$ 0.88 & 84.12 $\pm$ 2.75\\
     \multicolumn{1}{c|}{} & \textbf{\proj-edge} &  \underline{98.20 $\pm$ 0.32} & {99.30 $\pm$ 0.07}  &  \underline{97.95 $\pm$ 0.41} & \textbf{98.86 $\pm$ 0.03} & \textbf{86.16 $\pm$ 0.76} & \underline{90.54 $\pm$ 1.05} \\
     \multicolumn{1}{c|}{} & \textbf{\proj-node} &  \underline{98.00 $\pm$ 0.47} & {99.18 $\pm$ 0.10} & 96.81 $\pm$ 0.56 & \textbf{98.78 $\pm$ 0.13}  &  \underline{84.41 $\pm$ 0.83} & \underline{90.82 $\pm$ 1.40} \\
    \hline
    \end{tabular}
    }
    \caption{\small Link prediction performance in  average precision (AP) (mean in percentage $\pm$ 95$\%$ confidence level). \textbf{Bold font} and {underline} highlight the best performance and the second best performance on average.}
    \label{apx:ap}
\end{table*}

\begin{table*}[t]
    \centering
    \resizebox{0.7\textwidth}{!}{%
    \begin{tabular}{c|l|ccccc}
    \hline
    \multicolumn{1}{c|}{Task} & Method & \multicolumn{1}{c}{Wikipedia} & \multicolumn{1}{c}{Reddit}& \multicolumn{1}{c}{GDELT}  &  \multicolumn{1}{c}{Ubuntu} & \multicolumn{1}{c}{Wiki-talk}\\
    \hline
    \multirow{10}{*}{\rotatebox[origin=c]{90}
    {Transductive}}
    & TGN-trunc & \underline{60.28 $\pm$ 4.09} & \underline{74.38 $\pm$ 1.72} & {75.78 $\pm$ 8.67}   & {20.94 $\pm$ 6.72} & {31.00 $\pm$ 5.02}\\
     & TGN-unif &  {52.05 $\pm$ 4.65} & {65.36 $\pm$ 4.49} & {71.78 $\pm$ 0.16}  & {21.87 $\pm$ 3.60} & {22.72 $\pm$ 6.55}\\
      & TGAT-trunc &  {45.91 $\pm$ 1.15} & {60.19 $\pm$ 1.05} & \underline{79.11 $\pm$ 0.04}  & { 22.37 $\pm$ 0.97} & {30.01 $\pm$ 0.84}\\
       & TGAT-unif &  {23.17 $\pm$ 0.46} & {45.90 $\pm$ 0.47} & {74.89 $\pm$ 0.05} & {16.61 $\pm$ 0.44} & {18.01 $\pm$ 0.43}\\
        & APAN-trunc &  {26.51 $\pm$ 1.46} & {43.68 $\pm$ 5.49} & {74.49 $\pm$ 4.14} & {5.79 $\pm$ 1.81} & {18.91 $\pm$ 9.17}\\
         & APAN-unif &  {30.59 $\pm$ 3.51} & {36.29 $\pm$ 10.12} & {72.53 $\pm$ 0.22} & {11.60 $\pm$ 4.22} & {9.99 $\pm$ 1.77}\\
           & NAT & \textbf{77.51 $\pm$ 22.83} & \textbf{90.40 $\pm$ 4.63} & {GPU OOM}  & \textbf{46.38 $\pm$ 9.14} & \textbf{74.17 $\pm$ 28.67}\\
    & NAT-node &  {39.58 $\pm$ 6.21} & {39.89 $\pm$ 8.27} & {GPU OOM} & {18.93 $\pm$ 3.10} & {21.32 $\pm$ 3.59}\\
     \multicolumn{1}{c|}{} & \textbf{\proj-edge} &  \underline{61.90 $\pm$ 2.31} & \underline{75.40 $\pm$ 1.85} & \textbf{84.45 $\pm$ 0.84} & {24.34 $\pm$ 1.89} & \underline{32.57 $\pm$ 1.49} \\
     \multicolumn{1}{c|}{} & \textbf{\proj-node} & {54.42 $\pm$ 3.08} & {67.65 $\pm$ 5.31} & {76.60 $\pm$ 8.43}  & \underline{25.43 $\pm$ 0.97} & \underline{31.95 $\pm$ 2.64}\\
    \hline
    \multicolumn{1}{c|}{\multirow{10}{*}{\rotatebox[origin=c]{90}
    {Inductive}}}
    & TGN-trunc &   \underline{54.29 $\pm$ 3.13} & {58.20  $\pm$ 7.35} & {74.37 $\pm$ 1.26}  & { 19.58 $\pm$ 2.99} & \underline{27.93 $\pm$ 2.94}\\
     & TGN-unif &  {44.26 $\pm$ 2.32} & {53.02 $\pm$ 2.04} & \textbf{81.69 $\pm$ 3.73} & {18.14 $\pm$ 9.87} & {25.84 $\pm$ 0.68}\\
  & TGAT-trunc &  {45.18 $\pm$ 1.56} & {50.67 $\pm$ 5.39} & {59.90 $\pm$ 0.59} & \underline{22.41 $\pm$ 0.58} & {26.15 $\pm$ 1.50} \\
   & TGAT-unif &  {15.60  $\pm$ 1.39} & {21.14 $\pm$  6.09} & {51.72 $\pm$ 1.06}  & {12.02 $\pm$ 1.03} & {20.51 $\pm$ 1.01}\\
    & APAN-trunc &  {18.92 $\pm$ 0.28} & { 34.85 $\pm$ 4.38} & \underline{77.60 $\pm$ 0.23}  & {7.16 $\pm$ 1.01} & {15.56 $\pm$ 3.52}\\
     & APAN-unif &  {19.83 $\pm$ 1.77} & {38.15 $\pm$ 4.45} & {49.87 $\pm$ 3.95}  & {11.63 $\pm$ 9.75} & {17.46 $\pm$ 5.45} \\
       & NAT &  \textbf{75.02 $\pm$ 22.01} & \textbf{89.53 $\pm$ 4.08} & {GPU OOM}  & \textbf{37.81 $\pm$ 12.40} & \textbf{39.02 $\pm$ 24.29}\\
       & NAT-node &  {34.56 $\pm$ 4.18} & {50.58 $\pm$ 9.29} & {GPU OOM} & {19.64 $\pm$ 4.43} & {19.60 $\pm$ 6.73}\\
     \multicolumn{1}{c|}{} & \textbf{\proj-edge} &  {49.94 $\pm$ 2.12} & \underline{61.74  $\pm$ 1.85} & \textbf{81.60 $\pm$ 4.17} & {21.13 $\pm$ 2.44}  & {24.26 $\pm$ 4.60}  \\
     \multicolumn{1}{c|}{} & \textbf{\proj-node} &   {44.59 $\pm$ 6.36} & {53.43 $\pm$ 8.03} & {71.58 $\pm$ 14.65}  & {21.46 $\pm$ 2.97} & {25.39 $\pm$ 1.87} \\
    \hline
    \end{tabular}
    }
    \caption{\small Link prediction performance in  Mean Reciprocal Rank (MRR) with large number of negative samples per positive sample (mean in percentage $\pm$ 95$\%$ confidence level). For the largest-scale datasets GDELT and MAG, we use 50 negative samples while for other datasets we use 500. MAG is not evaluated because it takes a significantly long time to run all models with a large number of negative samples. \textbf{Bold font} and {underline} highlight the best performance and the second best performance on average.}
    \label{apx:mrr}
\end{table*}

The following are the detailed descriptions of the six datasets we tested.
\begin{itemize}
\item Wikipedia\footnote{\url{http://snap.stanford.edu/jodie/wikipedia.csv}}~\citep{WIKI, kumar2019predicting} records edit events on wiki pages. It is a bipartite graph where a set of nodes represents the editors and another set represents the wiki pages. The stamped link represents the edit events. The edge features are extracted from the contents of wiki pages. The node labels are binary which  indicates whether a user
is banned from posting. The original data dump is under the Creative Commons Attribution-Share-Alike 3.0 License and the Terms of Use is on the website (\url{https://dumps.wikimedia.org/legal.html}). The derived dataset by~\citet{kumar2019predicting} is open-sourced \href{http://snap.stanford.edu/jodie/wikipedia.csv}{here}.

\item Reddit\footnote{ \url{http://snap.stanford.edu/jodie/reddit.csv}}~\citep{REDDIT, kumar2019predicting} is a dataset of the post events by users on subreddits. It is also an attributed bipartite graph between users and subreddits. The node labels are also binary values that indicate whether a user is banned from posting to a subreddit. The original dataset has the Terms of Use in the website (\url{https://pushshift.io/signup}). The derived dataset by~\citet{kumar2019predicting} is open-sourced \href{http://snap.stanford.edu/jodie/reddit.csv}{here}.

\item GDELT\footnote{\url{https://github.com/tedzhouhk/TGL}}~\citep{zhou2022tgl} is a Temporal Knowledge Graph
(TKG) originated from the GDELT Event Database~\citep{GDELT} and adapted by TGL for richer node and edge features. It is a large-scale dataset with more than 100M links. It records events from news and articles in over 100 languages every 15 minutes. The nodes represents actors and the edges represent events that happen between a pair of actors. The node features represent the CAMEO codes of the actors and the link features represent the CAMEO codes of the events. The labels are the countries where the actors were located when the events happen. According to the Terms of Use in the website (\url{https://www.gdeltproject.org/}), it is an open platform for research and analysis of global society and thus all datasets released by the GDELT Project are available for unlimited and unrestricted use for any academic, commercial, or governmental use of any kind without fee. The derived dataset by~\citet{zhou2022tgl} is open-sourced \href{https://github.com/tedzhouhk/TGL}{here}.

MAG\footnote{\url{https://github.com/tedzhouhk/TGL}}~\citet{zhou2022tgl} is a homogeneous sub-graph of the
heterogeneous MAG240M graph in OGB-LSC~\citep{Hu2021OGB} extracted by TGL.  It is another large-scale dataset with more than 100 million nodes and 1.3 billion links. It is a
paper-paper citation network where each node in MAG represents
one academic paper. The temporal edges represent citation of one paper to another with
timestamp representing the year when the paper is published. The
node features are the embeddings of
the abstract of the paper generated with RoBERTa~\citep{Liu2019RoBERTa}. The node labels are the arXiv subject areas. The original dataset is licensed ODC-BY License.
 The derived dataset by~\citet{zhou2022tgl} is open-sourced \href{https://github.com/tedzhouhk/TGL}{here}.

\item Ubuntu\footnote{\url{https://snap.stanford.edu/data/sx-askubuntu.html}}~\citep{UBUNTU, paranjape2017motifs} is a dataset that records the events on a stack exchange web site called Ask Ubuntu.\footnote{\url{http://askubuntu.com/}} The nodes represent users and there are three different types of edges, (1) user $u$ answering user $v$'s question, (2) user $u$ commenting on user $v$'s question, and (3) user $w$ commenting on user $u$'s answer. The original data dump is cc-by-sa 4.0 licensed. The derived dataset by~\citet{paranjape2017motifs} is open-sourced \href{https://snap.stanford.edu/data/sx-askubuntu.html}{here}.

\item Wiki-talk\footnote{\url{https://snap.stanford.edu/data/wiki-talk-temporal.html}}~\citep{WIKITALK, paranjape2017motifs} is a dataset that represents the edit events on Wikipedia user talk pages. The dataset spans approximately 5 years so it accumulates a large number of nodes and edges. This is a large-scale dataset with more than 1M nodes. The original data dump is under the Creative Commons Attribution-Share-Alike 3.0 License. The derived dataset by~\citet{paranjape2017motifs} is open-sourced \href{https://snap.stanford.edu/data/wiki-talk-temporal.html}{here}.
\end{itemize}

\begin{table*}[t]
\centering \small
\resizebox{0.6\textwidth}{!}{%
\begin{tabular}{c|ccccccc}
\hline
\textbf{Params}   & \textbf{Wikipedia} & \textbf{Reddit}  & \textbf{GDELT} & \textbf{MAG} & \textbf{Ubuntu} & \textbf{Wiki-talk} \\ \hline
$s$         &  20 & 20 & 10          & 2 & 20 & 20   \\
\hline
\end{tabular}%
}
\caption{\small Number of sampled temporal neighbors for each dataset for all methods. GDELT and MAG use smaller $s$ because of CPU and GPU size limits.}
\label{tab:hyperparam}
\end{table*}

\begin{table*}[t]
\centering \small
\resizebox{0.6\textwidth}{!}{%
\begin{tabular}{c|ccccccc}
\hline
\textbf{Params}   & \textbf{Wikipedia} & \textbf{Reddit}  & \textbf{GDELT} & \textbf{MAG} & \textbf{Ubuntu} & \textbf{Wiki-talk} \\ \hline
Batch size     &  100 & 100 & 5000 & 5000 & 600 & 1000 \\
\hline
\end{tabular}%
}
\caption{\small The batch sizes used for scalability evaluation for each dataset for all methods.}
\label{tab:bs}
\end{table*}

\section{Baselines, hyperparameters and the experiment setup}\label{apd:hyper}

\textbf{TGAT}~\citep{xu2020inductive} adapts GAT~\cite{velivckovic2018graph} for static graphs to dynamic graphs. It aggregates messages from dynamic neighbors via an attention mechanism.   While it proposed recent sampling, its implementation is significantly inefficient because it has to calculate the recency weights online before sampling. Thus, we only consider uniform sampling and truncation. We use 2 attention heads and 100 hidden dimensions. We only consider aggregation from the first-hop neighbors because aggregating second-hop neighbors introduces even more severe latency for sampling. We also do not observe significant improvement in prediction performance using second-hop neighbors. We do not find an official licensed original implementation for this work and we adopt the implementation by TGL~\citep{zhou2022tgl}.

\textbf{TGN}~\cite{tgn_icml_grl2020}  is a recently proposed SOTA temporal graph representation learning approach. It keeps track of a memory state for each node and update with new interactions. We train TGN with 100 dimensions for each of memory module, time feature and node embedding. We also only consider sampling the first-hop neighbors because of computational efficiency. Their source code\footnote{\url{https://github.com/twitter-research/tgn}} is licensed under the Apache-2.0 License but we adopt the implementation by TGL.

\textbf{APAN}~\citep{wang2021apan} is a recent approach designed for low inference latency. It prepares the neighborhood information forward with graph propagation instead of aggregation. While it inspires our approach, it does not support recent sampling and has shown to consume significant CPU memory while performing subpar to SOTA methods. The reason is it stores the messages for each node in CPU which consumes massive CPU memory.  We train APAN with 100 dimensions for the node embeddings and 10 dimensions for the mailbox. Their original implementation\footnote{\url{https://github.com/WangXuhongCN/APAN}} is licensed under the MIT License but we adopt the implementation by TGL.

\textbf{TGL}~\citep{zhou2022tgl} is a very recent framework that support efficient and scalable temporal graph representation learning. It proposes temporal neighbor sampling with multi-core CPUs in parallel. Overall, it is shown to achieve on average 13$\times$ speedup for training and 173$\times$ speedup on neighbor sampling. For all of the baselines above, we evaluate them using the implementation by TGL. For the GDELT and the MAG datasets, since the node and edge features do not fit in our GPU, we set the \texttt{all\_on\_gpu} flag to \texttt{false}. For all other datasets, we set it to \texttt{true}. This work is licensed under the Creative Commons BY-NC-ND 4.0 International
License. The source code licensed under the Apache-2.0 License is provided \href{https://github.com/tedzhouhk/TGL}{here}.\footnote{\url{https://github.com/tedzhouhk/TGL}}

\textbf{NAT}~\citep{luo2022neighborhood} is also a very recent work that achieves SOTA link prediction performance for dynamic graphs. It proposed dictionary-type node representations to replace the traditional single-vector node representations which keep track of historical neighborhood information and avoid the expensive backward neighbor sampling. However, it relies on the construction of joint neighborhood features which only work with node pairs and cannot be easily generalized to node classification. Therefore, we evaluate it on link prediction only. We adapted NAT for node classification (called \textbf{NAT-node}) by removing the joint structural features and only aggregating their neighborhood representations to generate node representations. We also only use the first-hop neighborhood. Other than that, we follow the hyperparameter settings documented in their paper. Their implementation\footnote{\url{https://github.com/Graph-COM/Neighborhood-Aware-Temporal-Network}} is licensed under the MIT License.

\textbf{\proj} (our work). For both \proj-node and \proj-edge, we tune the node representation and the node status dimensions between 50 to 100. We use two attention heads and we tune $\alpha$ and dropout between 0 to 1 with grid search.

Lastly, we note that the Adam optimizer~\citep{adam} is used for all baselines.



\subsection{Inductive evaluation of \proj}\label{apd:inductive}
Our evaluation pipeline for inductive learning is similar to NAT~\citep{luo2022neighborhood} and different from other baselines. For backward sampling methods such as TGN~\cite{tgn_icml_grl2020} and TGAT~\cite{xu2020inductive}, when they do inductive evaluations, they can access the entire training and evaluation data for aggregation, including events that are masked for inductive test. However, \proj's forward sampling prepares the down-sampled neighbors forward for future use and ignores the masked events during the training phase. Thus, by the end of the training, even all historical events become accessible, \proj cannot leverage them. Therefore, to ensure a fair comparison, 
after training, \proj processes the \textbf{full} train and validation data  with all nodes unmasked to gather the down-sampled temporal neighbors from the complete set, and then processes the test data. Note that in this last pass over the \textbf{full} train and validation data, we only perform the forward sampling and do not perform training anymore.


\subsection{Node classification preprocessing}\label{apd:node_classification}
For node classification, we pre-process and combine the node labels to links so that each link can have a source label, a target label or both.  For each label $y_u^t$, we assign the label to the first link of $u$ that appears since time $t$. If $u$ is a source node, the label is assigned to source label and vice versa.  While processing the link stream, if a link has source or target labels, we record the node representations of the source or target node at that time and generate a prediction for the node classes.

\section{Evaluation on TGB}\label{apx:tgb}
TGB~\citep{huang2023temporal} is adopted by a lot of TGRL methods for standardizing the evaluation. Our main evaluation does not use TGB because it is insufficient for computational evaluation which involves billion-scale graphs. We provide some additional experiments for \proj using TGB here in Table~\ref{tab:tgb_results}. It verifies that \proj can match or outperform TGN in link prediction.

\begin{figure}
    \centering
    \resizebox{0.4\textwidth}{!}{%
    \begin{tabular}{r|ccc}
    \hline
    \multicolumn{1}{r}{Method}  & \multicolumn{1}{c}{tgbl-review} & \multicolumn{1}{c}{tgbl-coin}&\multicolumn{1}{c}{tgbl-comment}\\
    \hline
    
     TGN &  {34.90 $\pm$ 2.00} &  {58.60 $\pm$ 3.70} & {37.90 $\pm$ 2.10} \\
     

     DyRep & {22.00 $\pm$ 3.00} &  {45.20 $\pm$ 4.60} &  {28.90 $\pm$ 3.30}\\
      
      EdgeBank(tw) & {2.50} &  {58.00} &  
14.90 \\
     \textbf{\proj-edge} & {35.08 $\pm$ 2.20} & {61.92 $\pm$ 1.43} & {38.84 $\pm$ 2.17}  \\
      \textbf{\proj-node} & 37.13 $\pm$ 0.40 & 58.66 $\pm$ 2.14 & 36.91 $\pm$ 0.71 \\
    \hline
    \end{tabular}
    }
    \captionof{table}{\small Link prediction performance in Mean Reciprocal Rank (MRR) evaluated on TGB.} 
    
    \label{tab:tgb_results}
\end{figure}

\section{TGL Node Classification for MAG}\label{apx:tgl_mag}
We fail to reproduce the node classification scores for MAG over all of the baselines implemented in TGL~\citep{zhou2022tgl}. Our hypothesis is as follows. TGL has reported the scores for MAG based on an implementation with an issue. For the backtracking and sampling of temporal neighbors to learn representation at time $t$, TGL considers temporal neighbors of the same timestamp $t$ which may cause information leakage. We noticed this issue in their multiple-gpu training script. We evaluate TGL using their single-gpu training script which does not have the issue and results in the worse prediction for MAG as shown in Table~\ref{tab:auc node results}.

\begin{figure}[t]
    \centering
    \includegraphics[trim={0.3cm 0.3cm 0.3cm 0.3cm}, clip, width=0.5\textwidth]{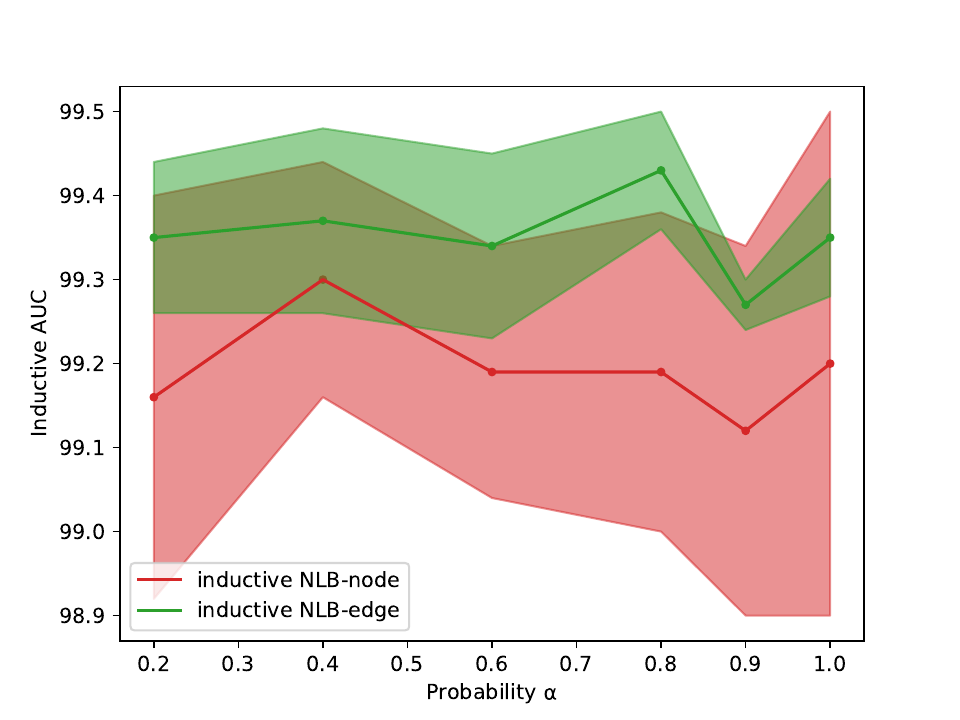}
    \vspace{-0.1cm}
    \caption{\small{The changes in inductive link prediction test performance with respect to different $\alpha$ on the Reddit dataset.}}
    \label{fig:alpha_reddit_inductive}
    \vspace{-3mm}
\end{figure}
\begin{figure}[t]
    \centering
    \includegraphics[trim={0.3cm 0.3cm 0.3cm 0.3cm}, clip, width=0.5\textwidth]{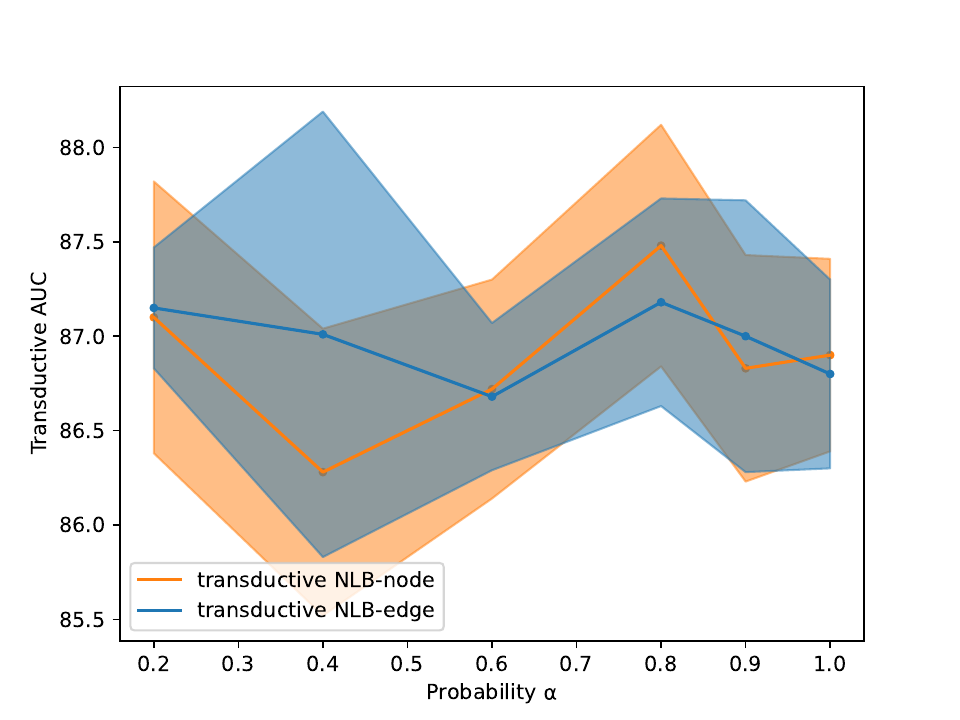}
    \vspace{-0.1cm}
    \caption{\small{The changes in transductive link prediction test performance with respect to different $\alpha$ on the Ubuntu dataset.}}
    \label{fig:alpha_ubuntu_transductive}
    \vspace{-3mm}
\end{figure}

\section{The large difference between CPU and GPU energy consumption}
\label{apd:energy}
While we directly obtained the numbers from the PyJoules\footnote{\url{https://pyjoules.readthedocs.io/en/latest/}} output, we observe a large difference between CPU and GPU energy consumption. We find that it can be attributed to the following factors:

(1) Both the CPU and GPU have default energy costs even when no applications are running, as there may be background applications that constantly require energy. Within 1 minute, the default energy cost for the CPU is roughly 9,000 MJ, while the GPU cost is ~0.8MJ.

(2) The training process is notably more energy-intensive on the CPU compared to the GPU, as there can be a considerable amount of CPU-GPU communication that is also counted on the CPU cost side as CPU controls the communication. This difference can be even more significant for applications that sample on the CPU side. For the NLB, we observed an additional ~2,500MJ CPU energy and ~12MJ GPU energy for training on the Wikitalk dataset. For TGN-uniform, it requires ~10,000MJ CPU energy and ~10MJ GPU energy in extra.

\section{The efficiency evaluation of different sampling methods}
\label{apd:sampling}
In this section, we give additional details to show that forward sampling is indeed more efficient than samplings used by the conventional TGRL methods in practice. During the inference phase, both the proposed forward sampling and truncation methods can retrieve the down-sampled neighbors of a node in $O(1)$ time, while uniform sampling takes $O(N_u^t)$ time, where $N_u^t$ is the number of neighbors of node u at time t. Since we typically need to infer a batch of nodes at a time, we prefer the sampling for the entire batch to be conducted in parallel. Both truncation and uniform sampling require storing and sampling from historical interactions. Due to the uneven distribution of degrees among different nodes, these operations can only be carried out within the CPUs, even when performed in parallel. On the contrary, our method can directly retrieve the already sampled neighbors within the GPU in constant time.

We have conducted an experiment that measures only the accumulated sampling time during inference for one epoch in seconds for three datasets (Table~\ref{tab:samping}). The results show that our method can retrieve the samples instantaneously, while both truncation and uniform sampling are significantly slower in comparison. Uniform sampling is even slower than truncation, and multi-thread (32 threads) sampling is faster than single-thread sampling, especially for large datasets, as expected.

\begin{table*}[t]
    \centering     
\resizebox{0.85\textwidth}{!}{%
    \begin{tabular}{r|ccccc}
    \hline
    \multicolumn{1}{r}{dataset}  & \multicolumn{1}{c}{proposed sampling} & \multicolumn{1}{c}{trunc.~1-thread}&\multicolumn{1}{c}{trunc.~32-thread} & \multicolumn{1}{c}{unif.~1-thread} & \multicolumn{1}{c}{unif.~32-thread}\\
    \hline
    
     Wiki-talk &  0.19 &  6.06 & 5.47  & 8.90 & 6.53 \\
      Reddit &  0.10 & 0.96 & 0.68 & 1.02 & 0.82 \\
      GDELT &  0.55 & 119.19 & 93.46 & 390.24 & 181.47

 \\
    \hline
    \end{tabular}
    }
    \vspace{-1mm}
    \caption{\small The accumulated sampling time for inference during the training process over one epoch in seconds.}
    \label{tab:samping}
\end{table*}

\begin{table*}[t]
    \centering     
\resizebox{1.0\textwidth}{!}{%
    \begin{tabular}{r|c|c|cccccc}
    \hline
    \multicolumn{1}{r}{} & \multicolumn{1}{c}{task} & \multicolumn{1}{c}{dataset} & \multicolumn{1}{c}{0.2}&\multicolumn{1}{c}{0.4} & \multicolumn{1}{c}{0.6} & \multicolumn{1}{c}{0.8} & \multicolumn{1}{c}{\underline{0.9}} & \multicolumn{1}{c}{1.0}\\
    \hline
    
     NLB-edge &  \multirow{2}{*}{trans.} &
\multirow{2}{*}{Reddit} &  $72.39  \pm 2.15$ & $72.86  \pm 2.06$ & $73.13  \pm 3.44$ & $73.47  \pm 2.48$ & $75.40 \pm 1.85$ & $72.77 \pm 2.69$ 
\\
      NLB-node   
& & &  $66.21 \pm 2.95$ & $64.85 \pm 3.83$ & $67.27 \pm 2.00$ & $68.15 \pm 1.45$ & $67.65 \pm 5.31$ & $67.85 \pm 2.72$

 \\
 \hline
    NLB-edge & \multirow{2}{*}{induc.} & \multirow{2}{*}{Ubuntu}  & 
 $21.59 \pm 4.83$ & $21.74 \pm 2.84$ & $22.72 \pm 3.35$ & $23.84 \pm 2.27$ & $21.13 \pm 2.44 $ & $21.49 \pm 4.49$ 
 \\
 NLB-node &  & &
 $22.73 \pm 3.07$ & $21.08 \pm 3.01$ & $20.60 \pm 4.22$ & $20.54 \pm 3.48$ & $21.46 \pm 2.97$ & $21.30 \pm 4.92$
 \\
    \hline
    \end{tabular}
    }
    \vspace{-1mm}
    \caption{\small The performance measured in MRR with 500 negative samples of NLB-edge and NLB-node given different $\alpha$'s. $\alpha = 0.9$
is the setting used for comparison with baselines.}
\label{MRR_alpha}
\end{table*}

\begin{table*}[t]
    \centering
\resizebox{1.0\textwidth}{!}{%
    \begin{tabular}{r|c|ccccccc}
    \hline
    \multicolumn{1}{r}{}  & 
\multicolumn{1}{c}{task} &\multicolumn{1}{c}{0} & \multicolumn{1}{c}{5}&\multicolumn{1}{c}{10} & \multicolumn{1}{c}{15} & \multicolumn{1}{c}{\underline{20}} & \multicolumn{1}{c}{25} & \multicolumn{1}{c}{30}\\
    \hline
    
     NLB-edge & \multirow{2}{*}{trans.} & 
$62.35 \pm 1.30$ & $70.62\pm 3.16$ & $69.93 \pm 4.06$ & $73.13 \pm 1.98$ & $75.40 \pm 1.85$ & $73.52 \pm 1.34$ & $72.30 \pm 3.38$
\\
      NLB-node   & 
& $62.35 \pm 1.30$ & $66.14 \pm 4.74$ & $68.96 \pm 1.67$ & $69.17 \pm 1.82$ & $67.65 \pm 5.31$ & $67.10 \pm 4.42$ & $66.78 \pm 1.94$ 
 \\
 \hline
    NLB-edge & \multirow{2}{*}{induc.} & $36.59 \pm 3.49$ & $54.65\pm 4.36$ & $59.62 \pm 1.03$ & $56.09 \pm 5.45$ & $61.74\pm 1.85$ & $61.87 \pm 2.03$ & $61.33 \pm 2.65$ 
 \\
 NLB-node & & $36.59 \pm 3.49$ & $47.69 \pm 6.44$ & $51.15 \pm 5.36$ & $54.50 \pm 4.76$ & $53.43 \pm8.03$ & $57.06 \pm 7.02$ & $58.09 \pm 0.93$ 
 \\
    \hline
    \end{tabular}
    }
    \vspace{-1mm}
    \caption{\small The performance measured in MRR with 500 negative samples of NLB-edge and NLB-node on the Reddit dataset given different down-sampled neighbor hash table sizes $s$. $s = 20$
is the setting used for comparison with baselines.}
\label{MRR_s}
\end{table*}

\end{document}